\documentclass{article}

\usepackage{microtype}
\usepackage{graphicx}
\usepackage{subfigure}
\usepackage{booktabs} % for professional tables

\usepackage{hyperref}
\usepackage{etoolbox}
\makeatletter
\patchcmd\@combinedblfloats{\box\@outputbox}{\unvbox\@outputbox}{}{\errmessage{\noexpand patch failed}}
\makeatother

\usepackage[utf8]{inputenc} % allow utf-8 input
\usepackage[T1]{fontenc}    % use 8-bit T1 fonts
\usepackage{url}            % simple URL typesetting
\usepackage{amsfonts}       % blackboard math symbols
\usepackage{nicefrac}       % compact symbols for 1/2, etc.
\usepackage{amsmath, amsthm, amssymb}
\newtheorem{prop}{Proposition}
\usepackage{xcolor}
\usepackage{bm}
\newcommand{\bA}{\bm A}
\newcommand{\bX}{\bm X}
\newcommand{\bx}{\bm x}

\newcommand{\bz}{\bm z}

\newcommand{\bS}{\bm S}
\newcommand{\bs}{\bm s}

\newcommand{\bU}{\bm U}
\newcommand{\bV}{\bm V}
\newcommand{\brho}{\bm \rho}
\newcommand{\btau}{\bm \tau}
\newcommand{\bphi}{\bm \phi}
\newcommand{\bdelta}{\bm \delta}
\newcommand{\bxi}{\bm \xi}
\newcommand{\bzero}{\bm 0}
\newcommand{\bOne}{\bm 1}

\newcommand{\bSigma}{\bm \Sigma}
\newcommand{\bOmega}{\bm \Omega}
\newcommand{\bomega}{\bm \omega}
\newcommand{\bTheta}{\bm \Theta}
\newcommand{\btheta}{\bm \theta}

\newcommand{\logdet}{\log\mbox{det }}

\usepackage[accepted]{icml2019}

\icmltitlerunning{Bayesian Joint Spike-and-Slab Graphical Lasso}

\begin{document}

\twocolumn[
\icmltitle{Bayesian Joint Spike-and-Slab Graphical Lasso}



\icmlsetsymbol{equal}{*}

\begin{icmlauthorlist}
\icmlauthor{Zehang Richard Li}{yale}
\icmlauthor{Tyler H. McCormick}{uw1,uw2}
\icmlauthor{Samuel J. Clark}{osu}
\end{icmlauthorlist}

\icmlaffiliation{yale}{Department of Biostatistics, Yale School of Public Health, New Haven, Connecticut, USA}
\icmlaffiliation{uw1}{Department of Statistics, University of Washington, Seattle, Washington, USA}
\icmlaffiliation{uw2}{Department of Sociology, University of Washington, Seattle, Washington, USA}
\icmlaffiliation{osu}{Department of Sociology, Ohio State University, Columbus, Ohio, USA}

\icmlcorrespondingauthor{Zehang Richard Li}{zehang.li@yale.edu}

\icmlkeywords{graphical model, EM algorithm}

\vskip 0.3in
]



\printAffiliationsAndNotice{} % otherwise use the standard text.

% Place figure captions \emph{under} the figure (and omit titles from inside
%     the graphic file itself). Place table captions \emph{over} the table.
% \item References must include page numbers whenever possible and be as complete
%     as possible. Place multiple citations in chronological order.

\begin{abstract}
In this article, we propose a new class of priors for Bayesian inference with multiple Gaussian graphical models. We introduce Bayesian treatments of two popular procedures, the group graphical lasso and the fused graphical lasso, and extend them to a continuous spike-and-slab framework to allow self-adaptive shrinkage and model selection simultaneously.  We develop an EM algorithm that performs fast and dynamic explorations of posterior modes. Our approach selects sparse models efficiently and automatically with substantially smaller bias than would be induced by alternative regularization procedures. The performance of the proposed methods are demonstrated through simulation and two real data examples.
\end{abstract}

\section{Introduction}
\label{sec:intro}
Bayesian formulations of graphical models have been widely adopted as a way to characterize conditional independence structure among complex high-dimensional data.  These models are popular in scientific domains including genomics~\citep{briollais2016bayesian,Peterson2013}, public health~\citep{dobra2014graphical,li2017mix}, and economics~\citep{dobra2010modeling}. 
% One of the key advantage of the Bayesian approach in graphical modeling lies in its ability to incorporate prior information and natural extensions to hierarchical model structures. 
In practice, data often come from several distinct groups. For example, data may be collected under various conditions, at different locations and time periods, or correspond to distinct subpopulations. Assuming a single graphical model in such cases can lead to unreliable estimates of network structure, whereas the alternative, estimating different graphical models separately for each group, may not be feasible for high dimensional problems.

Several approaches have been proposed to learn graphical models jointly for multiple classes of data.  Much of this work extends the penalized maximum likelihood approach to incorporate additional penalty terms that encourage the class-specific precision matrices to be similar~\citep{guo2011joint, danaher2014joint,saegusa2016joint,ma2016joint}. In the Bayesian literature, \citet{Peterson2014} and \citet{lin2017joint} utilize Markov Random Field priors to model a super-graph linking different graphical models. \citet{tan2017bayesian} uses a logistic regression model to link the connectivity of nodes to covariates specific to each graph. These approaches only model the similarity of the underlying graphs, and thus are limited in their ability to borrow information when estimating the precision matrices.  Borrowing strength is especially important when some classes have small sample sizes.

In this work, we introduce a new Bayesian formulation for estimating multiple related Gaussian graphical models by leveraging similarities in the underlying sparse precision matrices directly. We first present two shrinkage priors for multiple related precision matrices, as the Bayesian counterpart of joint graphical lasso estimators~\citep{danaher2014joint}. We then propose a doubly spike-and-slab mixture extension to these priors, which allows us to achieve simultaneous shrinkage and model selection, as well as handle missing observations. In Section~\ref{sec:ecm} and \ref{sec:selection}, we extend from the recent literature on deterministic algorithms for Bayesian graphical models~\citep{gan2018bayesian,li2017ecm,deshpande2017simultaneous} and provide a fast Expectation-Maximization (EM) algorithm to quickly identify the posterior modes.  We also propose a procedure to sequentially explore a series of posterior modes. We then demonstrate the substantial improvements in both model selection and parameter estimation over the original joint graphical lasso approach using both simulated data and two real datasets in Section~\ref{sec:sim}. Finally, in Section~\ref{sec:discuss} we discuss future directions for improvements.

% Some authors have also encouraged to use a ridge penalty instead of lasso in estimating precision matrices when sample size is small~\citep{bilgrau2015targeted,van2016ridge,van2017mean}.  

\section{Preliminaries}
\label{sec:background}
\subsection{The joint graphical lasso}
\label{sec:prior}
We first briefly introduce the notation used throughout this paper. We let $G$ denote the number of classes in the data, and let $\bOmega_g$ and $\bSigma_g$ denote the precision and covariance matrix for the $g$-th class. We let $\omega^{(g)}_{jk}$ denote the $(j,k)$-th element in $\bOmega_g$ and $\bomega_{jk} = \{\omega^{(g)}_{jk}\}_{g = 1, ..., G}$ denote the vector of all the $(j,k)$-th elements in $\{\bOmega\}$. Suppose we are given $G$ datasets, $\bX^{(1)}, ..., \bX^{(G)}$, where $\bX^{(g)}$ is a $n_g \times p$ matrix of independent centered observations from the distribution $\mbox{Normal}(\bzero, \bOmega_g^{-1})$. As maximum likelihood estimates of $\bOmega_g$ can have high variance and are ill-defined when $p > n_g$, the joint penalized log likelihood for the $G$ dataset is usually considered instead:
\begin{equation}
\ell(\{\bOmega\}) = \frac{1}{2}\sum_{g=1}^{G} n_g \logdet \bOmega_g - tr(\bS_g\bOmega_g) - pen(\{\bOmega\}),
\label{eqn:penlikelihood}
\end{equation}
where $\bS_g = (\bX^{(g)})^T\bX^{(g)}$. The penalty function encourages $\{\bOmega\}$ to have zeros on the off-diagonal elements and be similar across groups. In particular, we consider two useful penalty functions studied in~\citet{danaher2014joint}, the group graphical lasso (GGL), and the fused graphical lasso (FGL):
% \begin{eqnarray}\label{eqn:ggl}
% \mbox{GGL:} \;\;\;\; pen(\{\bOmega\}) &=&\frac{\lambda_0}{2}\sum_{g}\sum_{j} |\omega_{jj}^{(g)}|+ \lambda_1\sum_{g}\sum_{j < k} |\omega_{jk}^{(g)}| + \lambda_2\sum_{j < k}||\bomega_{jk}||_2, \\\label{eqn:fgl}
% \mbox{FGL:} \;\;\;\; pen(\{\bOmega\}) &=& \frac{\lambda_0}{2}\sum_{g}\sum_{j} |\omega_{jj}^{(g)}| + \lambda_1\sum_{g}\sum_{j < k} |\omega_{jk}^{(g)}| + \lambda_2\sum_{j < k}\sum_{g<g'}|\omega_{jk}^{(g)} - \omega_{jk}^{(g')}|.
% \end{eqnarray}
\begin{eqnarray}\nonumber
pen(\{\bOmega\}) &=&\frac{\lambda_0}{2}\sum_{g}\sum_{j} |\omega_{jj}^{(g)}|+ \lambda_1\sum_{g}\sum_{j < k} |\omega_{jk}^{(g)}| \\\label{eqn:jgl}
&&+ \lambda_2\sum_{j < k}\widetilde{pen}(\bomega_{jk}), 
\end{eqnarray}
where $\widetilde{pen}(\bomega_{jk}) = ||\bomega_{jk}||_2$ for GGL and $\sum_{g<g'}|\omega_{jk}^{(g)} - \omega_{jk}^{(g')}|$ for FGL.
Both penalties encourage similarity across groups when $\lambda_2 > 0$, and reduce to separate graphical lasso problems when $\lambda_2 = 0$. The group graphical lasso encourages only similar patterns of zero elements across the $G$ precision matrices, while the fused graphical lasso encourages a stronger form of similarity: the values of off-diagonal elements are also encouraged to be similar across the $G$ precision matrices. In practice, $\lambda_0$ is typically set to $0$ when the diagonal elements are not to be penalized. 

\subsection{Bayesian formulation of Gaussian graphical models}
\label{sec:bayesGM}
One of the most popular approaches for Bayesian inference with Gaussian graphical models is the $G$-Wishart prior~\citep{lenkoski2011computational,Mohammadi2015a}. The $G$-Wishart prior estimates the precision matrices with exact zeros in the off-diagonal elements and enjoys the conjugacy with the Gaussian likelihood. However, posterior inference under the $G$-Wishart prior can be computationally burdensome and has to rely on stochastic search algorithms over the large model space, consisting of all possible graphs.
% The $G$-Wishart prior enjoys many  with the Gaussian likelihood and has been studied by many authors over the years. 
% Several authors have studied the Bayesian hierarchical models for joint estimation of multiple graphical models under the $G$-Wishart framework. Moreover, sampling from the $G$-Wishart distribution can pose computational challenges in high dimensions. 
In recent years, several classes of shrinkage priors have been proposed for estimating large precision matrices, including the graphical lasso prior~\citep{wang2012bayesian,Peterson2013}, the continuous spike-and-slab prior~\citep{wang2015scaling,li2017mix}, and the graphical horseshoe prior~\citep{li2017graphical}. This line of work draws direct connections between penalized likelihood schemes and, as their names suggest, the posterior modes in a Bayesian setting. 
Unlike the $G$-Wishart prior, these shrinkage priors do not take point mass at zero for the off-diagonal elements in the precision matrix, and thus usually lead to efficient block sampling algorithms with improved scalability. However, fully Bayesian procedures still need to rely on stochastic search to achieve model selection, making it less appealing for many problems.

To address this issue, deterministic algorithms have been proposed to perform fast posterior exploration and mode searching in Gaussian graphical models~\citep{gan2018bayesian,li2017ecm,deshpande2017simultaneous}. Motivated by the EMVS~\citep{rovckova2014emvs} and spike-and-slab lasso~\citep{Veronika2016spike} procedures in the linear regression literature, 
the idea is to use a two-component mixture distribution, i.e., spike-and-slab priors, to parameterize off-diagonal elements in the precision matrix, which allows simultaneous model selection and parameter estimation. 
% Moreover, due to the fast speed of such procedures, dynamic posterior exploration can be achieved by fixing the slab distribution and deploy the algorithm under a sequence of regularization parameter that governs the spike distribution. 
We will utilize a similar strategy for model estimation in this paper.

\section{Bayesian joint graphical lasso priors}
\label{sec:ssjgl}
We first provide a Bayesian interpretation of the group and fused graphical lasso estimators. From a probabilistic perspective, it is well understood that estimators that optimize a penalized likelihood can often be seen as the posterior mode estimator under some suitable prior distributions. The Bayesian counterpart to (\ref{eqn:jgl}) can be constructed by placing the prior $p(\{\bOmega\}) \propto \exp(-pen(\{\bOmega\}))$ on the precision matrices. Following directly from the Bayesian representation of lasso variants demonstrated in~\citet{Kyung2010}, we can rewrite $p(\{\bOmega\})$ as products of scale mixtures of normal distributions on the off-diagonal elements. That is, for the GGL prior, we can let

\begin{align}\nonumber
p(\{\bOmega\}|\btau, \brho) =&\; C_{\btau, \brho}^{-1} \prod_{j<k}\mbox{Normal}(\bomega_{jk}; \bzero, (\bTheta_{jk}^{(G)})^{-1})\\\label{eqn:ggl-1}
& \prod_{g}\prod_{j} \mbox{Exp}(\omega_{jj}^{(g)}; \frac{\lambda_0}{2}) \bOne_{\{\bOmega\} \in \{M^{+}\}},\\\label{eqn:ggl-2}
\bTheta_{jk}^{(G)} =&\; \mbox{diag}(\{\frac{1}{\rho_{jk}} + \frac{1}{\tau_{jkg}}\}_{g=1,...,G}),\\\nonumber
p( \btau, \brho) \propto&\; C_{\btau, \brho} \prod_{j<k} \Big(
\exp(-\frac{\lambda_1^2}{2}\sum_g \tau_{jkg} - \frac{\lambda_2^2}{2}\rho_{jk})
\rho_{jk}^{-\frac{1}{2}} \\\label{eqn:ggl-3}
& \prod_g(\tau_{jkg}(\frac{1}{\tau_{jkg}} + \frac{1}{\rho_{jk}}))^{-\frac{1}{2}} 
\Big),
\end{align} 

 where $C_{\btau, \rho}$ is a normalizing constant and $M^{+}$ denotes the space of symmetric positive definite matrices. The normalizing constant is analytically intractable due to this constraint, but it cancels out in the marginal distribution of $p(\{\bOmega\})$. Such cancellation has been studied by several authors~\citep{wang2012bayesian,wang2015scaling,liu2014bayesian}. Similarly, the FGL prior can be defined as
\begin{align}\nonumber
p(\{\bOmega\} | \btau, \bphi) =&\; C_{\btau, \bphi}^{-1} \prod_{j<k}\mbox{Normal}(\bomega_{jk}; \bzero, (\bTheta_{jk}^{(F)})^{-1})\\\label{eqn:fgl-1}
&\prod_{g}\prod_{j} \mbox{Exp}(\omega_{jj}^{(g)}; \frac{\lambda_0}{2}) \bOne_{\{\bOmega\} \in \{M^{+}\}},\\\label{eqn:fgl-2}
\bTheta_{jk}^{(F)} =&\; \begin{cases}
    \theta_{gg} = \frac{1}{\tau_{jkg}} + \sum_{g \neq g'} \frac{1}{\phi_{jkgg'}} & g = 1,...,G\\
    \theta_{gg'} = -\frac{1}{\phi_{jkgg'}} & g' \neq g
  \end{cases} \\\nonumber
  p(\btau, \bphi) \propto&\;
  C_{\btau, \bphi}  \prod_{j<k}
  \Big(
  |\bTheta_{jk}^{(F)}|^{-\frac{1}{2}}
\exp(-\frac{\lambda_1^2}{2}\sum_g \tau_{jkg} \\\label{eqn:fgl-3}
&- \frac{\lambda_2^2}{2}\sum_{g < g'}\phi_{jkgg'})
\prod_g \tau_{jkg}^{-\frac{1}{2}}
\prod_{g < g'} \phi_{jkgg'}^{-\frac{1}{2}}
\Big).
\end{align} 
It is also worth noting that both of the above priors are proper, and we leave the proof of the following proposition in the supplement.

\begin{prop}
The priors defined in (\ref{eqn:ggl-1}) -- (\ref{eqn:ggl-3}) and (\ref{eqn:fgl-1}) -- (\ref{eqn:fgl-3})  are proper and the posterior mode of $\{\bOmega\}$ is the solution of the group and fused graphical lasso problem with penalty terms defined in (\ref{eqn:jgl}).
\label{prop:proper}
\end{prop}

\section{Bayesian joint spike-and-slab graphical lasso priors}
\label{sec:dssjgl}
The Bayesian formulation of the joint graphical lasso problems discussed in the previous section provide shrinkage effects at the level of both individual precision matrices and across different classes. However, two issues remain. First, shrinkage priors alone do not produce sparse models since the posterior draws are never exactly $0$. Thus, additional thresholding is needed to obtain a sparse representation of the graph structure. Second, the fixed penalty term, $\lambda_1$ and $\lambda_2$ may be too restrictive, as the non-zero elements in $\{\bOmega\}$ are penalized equally to elements close to zero~\citep{li2017ecm}. To reduce the bias from over-penalizing the large elements, different hyper-priors on $\lambda_1$ have been proposed to adaptively estimate the penalty term in Bayesian graphical lasso~\citep{wang2012bayesian,Peterson2013}. 

Here we address both challenges simultaneously using the spike-and-slab approaches in Bayesian variable selection~\citep{george1993variable}. In particular, we employ a set of latent indicators to construct a ``selection'' prior on both the group level and within-groups for the similarity penalties. We first let binary variables $\bdelta = \{\delta_{jk}\}_{j<k}$ denote the existence of each edge in the graph, indexing the $2^{p(p-1)/2}$ possible models at the group level, so that $\delta_{jk}=1$ indicates the $(j,k)$-th edge is selected for all precision matrices. We then let another set of binary variables $\bxi = \{\xi_{jk}\}_{j<k}$ denote the \textit{non-existence} of `similarities' among the elements in the same cell of different precision matrices, so that $\xi_{jk}=0$ indicates the $(j,k)$-th element is expected to be similar. We use the term `similarity' here as a broad term parameterized by $\lambda_2$, since the behavior of the similarity depends on the form of the penalization.  
Conditional on the two binary indicators, we replace the fixed penalty parameters $\lambda_1$ and $\lambda_2$ by a mixture of edge-wise penalties that take values from $\{\lambda_1/v_0, \lambda_1/v_1\}$, and $\{\lambda_2/v_0, \lambda_2/v_1\}$ respectively, with fixed $v_1 > v_0 > 0$. 
That is, we introduce the following penalties conditional on $\bdelta$ and $\bxi$, and we propose the \textit{doubly spike-and-slab} extensions to GGL and FGL as
% \begin{align}\label{eqn:ss-ggl}
% \mbox{DSS-GGL:} \; pen(\{\bOmega\} | \bdelta, \bxi)&=\frac{\lambda_0}{2}\sum_{g}\sum_{j} |\omega_{jj}^{(g)}|+ \lambda_1\sum_{g}\sum_{j < k} \frac{|\omega_{jk}^{(g)}|}{v_{\delta_{jk}}} + \lambda_2\sum_{j < k}\frac{||\bomega_{jk}||_2}{v_{\xi^{*}_{jk}}}, \\\label{eqn:ss-fgl}
% \mbox{DSS-FGL:} \;pen(\{\bOmega\}| \bdelta, \bxi)&=\frac{\lambda_0}{2}\sum_{g}\sum_{j} |\omega_{jj}^{(g)}| + \lambda_1\sum_{g}\sum_{j < k} \frac{|\omega_{jk}^{(g)}|}{v_{\delta_{jk}}} + \lambda_2\sum_{j < k}\sum_{g<g'}\frac{|\omega_{jk}^{(g)} - \omega_{jk}^{(g')}|}{v_{\xi^{*}_{jk}}}.
% \end{align}
\begin{align}\nonumber
pen(\{\bOmega\} | \bdelta, \bxi)=&\;\frac{\lambda_0}{2}\sum_{g}\sum_{j} |\omega_{jj}^{(g)}|+ \lambda_1\sum_{g}\sum_{j < k} \frac{|\omega_{jk}^{(g)}|}{v_{\delta_{jk}}} \\\label{eqn:ss-jgl}
&+ \lambda_2\sum_{j < k}\frac{\widetilde{pen}(\bomega_{jk})}{v_{\xi^{*}_{jk}}},
\end{align}
where $\widetilde{pen}(\bomega_{jk})$ is defined as before and $\xi^{*}_{jk} = \xi_{jk} \delta_{jk}$. The prior defined in (\ref{eqn:ss-jgl}) relate to the unconditional penalties by 
$
pen(\{\bOmega\}) = pen(\{\bOmega\} | \bdelta, \bxi) - \log(p(\bdelta, \bxi))
$, and we will refer to them as  DSS-FGL and DSS-GGL.

In practice, we find it usually reasonable to enforce all elements from the spike distribution to also be similar, since the spike distribution is always chosen to have large penalization and leads to posterior modes at exactly $0$. However, other types of element-wise dependence between $\delta_{jk}$ and $\xi_{jk}$ are also possible with minor modifications. For example, we can also fix $\xi_{jk}$ to be $1$, so that the two penalty terms will always be proportional. We refer to this setting as spike-and-slab group and fused lasso (SS-GGL and SS-FGL) and discuss their behavior in the supplements. 

The original GGL and FGL suffer from the same bias induced by the excessive shrinkage of lasso estimates. With the introduction of $v_0$ and $v_1$, we can adaptively estimate which $\bomega_{jk}$ to penalize in a data-driven way. As we discuss in more detail in Section~\ref{sec:selection}, this adaptive shrinkage property can indeed significantly reduce bias imposed on the lasso penalty. That is, by choosing the hyperparameters so that $\lambda_i/v_0 \gg \lambda_i/v_1$, we impose only minimal shrinkage on values arising from the slab distribution.  From now on, in order to avoid confusion from the overparameterization, we always fix $v_1=1$, and report results with the effective shrinkage parameters $\lambda_{i}/v_{j}, i,j \in\{1,2\}$. At this point, it may still seem that we need introduced one more hyperparameter that needs to be tunned, but as we show in Section~\ref{sec:selection}, model selection can be achieved automatically without cross-validation.

For a Bayesian setup, we employ standard priors on the binary indicators to allow the edges to further share information on the sparsity level. The full generative model for $\{\bOmega\}$ is:
\begin{align}
\label{eqn:ss1}
% p(\{\bOmega\}|\bdelta) &\propto& C_{\btau, \rho}^{-1}C_{\bdelta}^{-1} \prod_{j<k}\mbox{Normal}(\bomega_{jk}; \bzero, v^2_{\delta_{jk}}\bTheta_{jk}^{-1})\prod_{g}\prod_{j} \mbox{Exp}(\omega_{jj}^{(g)}; \frac{\lambda_0}{2}) \bOne_{\{\bOmega\} \in \{M^{+}\}},\\
p(\{\bOmega\}|\bdelta,\bxi, \theta) =&\;  C_{\theta}^{-1}C_{\bdelta, \bxi}^{-1} \exp(-pen(\{\bOmega\} | \bdelta, \bxi))\bOne_{\{\bOmega\} \in \{M^{+}\}},\\
\nonumber
p(\bdelta, \bxi |\pi_{\bdelta}, \pi_{\bxi}) \propto&\; C_{\bdelta, \bxi} \prod_{j<k} \Big( \pi_{\bdelta}^{\delta_{jk}}(1-\pi_{\bdelta})^{1-\delta_{jk}}
\\\label{eqn:ss2}
&\pi_{\bxi}^{\xi_{jk}}(1-\pi_{\bxi})^{1-\xi_{jk}} \Big).
\end{align} 
where $\theta$ denote $(\btau, \brho)$ for DSS-GGL, and $(\btau, \bphi)$ for DSS-FGL. $C_{\bdelta, \bxi}$ is another intractable normalizing constant. We put standard Beta hyperpriors on the sparsity parameters so that $\pi_{\bdelta} \sim \mbox{Beta}(a_1, b_1)$ and $\pi_{\bxi} \sim \mbox{Beta}(a_2, b_2)$. Throughout this paper, we let $a_1=a_2=1$ and $b_1=b_2=p$.

Additionally, the above prior can be easily reparameterized with scale mixture of normal prior distributions similar as before by modifying the precision matrix $\bTheta$ into the following form, and they can be shown to be proper priors (the proofs can be found in the supplement):
% they are proper distributions with the following proposition. The proof can be found in the supplement.
\begin{eqnarray} \label{eqn:theta-ssfgl}
\tilde\bTheta_{jk}^{(F)} &=& \begin{cases}
    \theta_{gg} = \frac{v_{\delta_{jk}}}{\tau_{jkg}} + \sum_{g < g'} \frac{v_{\xi^{*}_{jk}}}{\phi_{jkgg'}} & g = 1,...,G\\
    \theta_{gg'} = -\frac{v_{\xi^{*}_{jk}}}{\phi_{jkgg'}} & g' \neq g
  \end{cases} \\\label{eqn:theta-ssggl}
\tilde\bTheta_{jk}^{(G)} &=& \mbox{diag}(\{\frac{v_{\xi^{*}_{jk}}}{\rho_{jk}} + \frac{v_{\delta_{jk}}}{\tau_{jkg}}\}_{g=1,...,G}).
\end{eqnarray}

\begin{prop}
The priors defined in (\ref{eqn:ss1}) -- (\ref{eqn:theta-ssggl}) are proper, and the posterior mode of $\{\bOmega\}$ is the solution to the corresponding spike-and-slab version of joint graphical lasso penalties.
\label{prop:proper2}
\end{prop}
 
Finally, it is straightforward to see that the proposed DSS-GGL and DSS-FGL penalties reduce to their non spike-and-slab counterparts when $\bdelta$ and $\bxi$ are fixed to be $1$. Several other spike-and-slab formulations in the literature can be seen as the special case of this prior when $G=1$ as well. For example, the spike-and-slab mixture of double exponential priors considered in~\citet{deshpande2017simultaneous} is a special case with $\lambda_2=0$. The spike-and-slab Gaussian mixtures in~\citet{li2017ecm} can also be considered as a special case where we further fix $\tau_{jkg} = \infty$. This approach is also related to the work on sparse group selection in linear regression, as has been discussed in~\citet{xu2015bayesian} and \citep{zhang2014bayesian}. As opposed to the point mass priors for the spike distribution commonly in the literature, our doubly spike-and-slab formulation of continuous mixtures allows the spike distribution to absorb small non-zero noises and facilitates fast dynamic explorations, as we will show in Section~\ref{sec:selection}. 
% We will refer to the Gaussian mixture case as group/fused spike-and-slab prior (GSS and FSS).
% For the selected edges, the penalty from $\lambda_2$ either further shrinks the elements to $0$ in $\bomega_{jk}$ under SS-GGL, or encourages them to be similar under SS-FGL.

% \section{Posterior inference via block Gibbs sampler}
% \label{sec:gibbs}

\section{Model estimation}
\label{sec:ecm}
Given fixed $\lambda_1$, $\lambda_2$, and $v_0$, The representation of $p(\{\bOmega\})$ with the scale mixture of normal distributions allows the posterior to be sampled using a block Gibbs algorithm,  
% where one column and row of the precision matrix in all classes are updated jointly condition on other parameters. 
as described in the supplement. However, choosing the hyperparameters can usually be a nontrivial task. Instead, we focus on faster deterministic methods to detect posterior modes under different choices of hyperparameters~\citep{rovckova2014emvs}. We present an EM algorithm that maximizes the complete-data posterior distribution $p(\{\bOmega\}, \bdelta, \bxi, \pi_{\bdelta}, \pi_{\bxi}|\bX)$ by treating the binary latent variables as ``missing data.'' Similar ideas have been explored in recent work for linear regression~\citep{rovckova2014emvs,Veronika2016spike} and single graphical model estimation~\citep{deshpande2017simultaneous,li2017ecm}. Our EM algorithm maximizes the objective function 
% $
% Q(\{\bOmega\}, \pi_{\bdelta}, \pi_{\bxi} | \{\bOmega\}^{(t)}, \pi_{\bdelta}^{(t)}, \pi_{\bxi}^{(t)}) = 
$E_{\bdelta, \bxi | \{\bOmega\}^{(t)}, \pi_{\bdelta}^{(t)}, \pi_{\bxi}^{(t)}, \bX}
(\log p(\{\bOmega\}, \pi_{\bdelta}, \pi_{\bxi} | \bX)|
% \{\bOmega\}^{(t)}, \pi_{\bdelta}^{(t)}, \pi_{\bxi}^{(t)}, \bX
\cdot
)
$
% in the $t$-th iteration.
% \begin{eqnarray*}
% Q(\{\bOmega\}, \pi_{\bdelta}, \pi_{\bxi} | \{\bOmega\}^{(t)}, \pi_{\bdelta}^{(t)}, \pi_{\bxi}^{(t)}) 
% &=& E_{\bdelta, \bxi | \{\bOmega\}^{(t)}, \pi_{\bdelta}^{(t)}, \pi_{\bxi}^{(t)}, \bX}
% (\log p(\{\bOmega\}, \pi_{\bdelta}, \pi_{\bxi} | \bX)|\{\bOmega\}^{(t)}, \pi_{\bdelta}^{(t)}, \pi_{\bxi}^{(t)}, \bX) \\
% &=& \mbox{constant} + \sum_g \frac{n_g}{2}\log|\bOmega_g| - \frac{1}{2}\sum_g tr(\bS_{g}\bOmega_g)
% - \frac{\lambda_0}{2}\sum_j\sum_g |\omega_{jj}^{(g)}|
% \\
% && - \lambda_1\sum_{j<k}\sum_g |\omega_{jk}^{(g)}|E_{\cdot|\cdot}[\frac{1}{v_0(1-\delta_{jk})+v_1\delta_{jk}}]
% + \sum_{j<k}\log(\frac{\pi_{\bdelta}}{1-\pi_{\bdelta}})E_{\cdot|\cdot}(\delta_{jk})
% \\
% && - \lambda_2\sum_{j<k} \mbox{pen}(\bomega_{jk})E_{\cdot|\cdot}[\frac{1}{v_0(1-\delta_{jk}\xi_{jk})+v_1\delta_{jk}\xi_{jk}}] 
% + \sum_{j<k}\log(\frac{\pi_{\bxi}}{1-\pi_{\bxi}})E_{\cdot|\cdot}(\xi_{jk})
% \\
% &&  + (a_1-1)\log(\pi_{\bdelta}) + (b_1 + \frac{p(p-1)}{2} - 1) \log(1-\pi_{\bdelta})\\
% && + (a_2-1)\log(\pi_{\bxi}) + (b_2 + \frac{p(p-1)}{2} - 1) \log(1-\pi_{\bxi}),
% \end{eqnarray*}
% where $E_{\cdot|\cdot}$ denotes conditional expectation $E_{\bdelta, \bxi | \{\bOmega\}^{(t)}, \pi_{\bdelta}^{(t)}, \pi_{\bxi}^{(t)}, \bX}$, and $\mbox{pen}(\bomega_{jk}) = ||\bomega_{jk}||_2$ for DSS-GGL and $\mbox{pen}(\bomega_{jk}) = \sum_{g<g'}|\omega_{jk}^{(g)} - \omega_{jk}^{(g')}|$ for DSS-FGL. 
% The EM algorithm proceeds 
by iterating between the E-step and M-step until changes in $\{\bOmega\}$ are within a small threshold. 

In the E-step, we compute the conditional expectation terms in the objective function. 
It turns out that it suffices to find the conditional distribution of $(\delta_{jk}, \xi_{jk})$. The corresponding cell probabilities are proportional the the following mixture densities:

\vspace{-1cm}
\begin{equation*}\hspace{-6cm}
	p^*_{\delta_{jk}, \xi_{jk}}(j,k) \propto	
\end{equation*}

\vspace{-2cm}
\begin{equation*}\label{eqn:pstar}
% p^*_{\delta_{jk}, \xi_{jk}}(j,k) \propto
\begin{cases}%\vspace{.2cm}
     \pi_{\bdelta}(1-\pi_{\bxi})\frac{\lambda_1\lambda_2}{v_0v_1}\psi(v_1, v_0)
     &\delta_{jk} = 1, \xi_{jk}=0 \vspace{.2cm}\\
    \pi_{\bdelta}\pi_{\bxi}\frac{\lambda_1\lambda_2}{v_1^2}\psi(v_1, v_1)
     &\delta_{jk} = 1, \xi_{jk}=1 \vspace{.2cm}\\
    (1-\pi_{\bdelta})(1-\pi_{\bxi})\frac{\lambda_1\lambda_2}{v_0^2}\psi(v_0, v_0)
     & \delta_{jk} = 0, \xi_{jk}=0
     % 0 & \delta_{jk} = 0, \xi_{jk}=1\\
   \end{cases}
\end{equation*}

where $\psi(a, b) = \exp(-\lambda_1\sum_g |\omega_{jk}^{(g)}|/a - \lambda_2 \widetilde{pen}(\bomega_{jk})/b)$.

It is interesting to note that the three scenarios above represent three types of relationships among $\bomega_{jk}$: weak shrinkage but strong similarity,  weak shrinkage and weak similarity, and strong shrinkage across classes. 
 $E_{\cdot|\cdot}(\delta_{jk})$ and $E_{\cdot|\cdot}(\xi_{jk})$ are then simply the marginal probabilities in this $2$ by $2$ table, i.e.,
$E_{\cdot|\cdot}(\delta_{jk}) = p^*_{1,0}(j,k) + p^*_{1,1}(j,k)$, and $E_{\cdot|\cdot}(\xi_{jk}) = E_{\cdot|\cdot}(\delta_{jk}\xi_{jk}) = p^*_{1,1}(j,k)$. 

The EM algorithm also handles missing cells in $\bX$ naturally. Assuming missing at random, the expectation can also be taken over the space of missing variables, by additionally computing
$
E_{\cdot|\cdot}(tr(\bS_g\bOmega_g)) = tr(
E_{\cdot|\cdot}((\bX^{(g)})^T\bX^{(g)})
\bOmega_g))
$, 
using the conditional Gaussian distribution of $\bm x_{i,miss}^{(g)}|\bm x_{i,obs}^{(g)}$. We relegate the derivations of the objective function to the supplement.
% \begin{equation}
% E_{\cdot|\cdot}((\bX^{(g)})^T\bX^{(g)}) = \sum_{i=1}^{n_g}E_{\cdot|\cdot}(\bx_i^{(g)})^TE_{\cdot|\cdot}(\bx_i^{(g)}) + 
% \end{equation}

Given the expectations calculated in the E-step, one might proceed with conditional maximization steps using gradient ascent similar to the Gibbs sampler~\citep{li2017ecm}. Alternatively, since the maximization step is equivalent to solving the following joint graphical lasso problem:
\begin{align*} \nonumber
\{\bOmega\} =&\; \mbox{argmax}_{\{\bOmega\}} \sum_g \frac{n_g}{2}\log|\bOmega_g| - \frac{1}{2}\sum_g tr(\bS_{g}\bOmega_g)\\
&- \frac{\lambda_0}{2}\sum_j\sum_g |\omega_{jj}^{(g)}| \\%\label{eqn:m-step}
&- \sum_{j<k}\lambda_1(\frac{p^*_{0,0}(j,k)}{v_0} + \frac{1-p^*_{0,0}(j,k)}{v_1})\sum_g |\omega_{jk}^{(g)}|\\
&- \sum_{j<k} \lambda_2(\frac{1-p^*_{1,1}(j,k)}{v_0} + \frac{p^*_{1,1}(j,k)}{v_1})\widetilde{pen}(\bomega_{jk}),
\end{align*}
meaning we can use the ADMM algorithm described in~\citet{danaher2014joint}.

\section{Dynamic posterior exploration}
\label{sec:selection}
\begin{figure*}[tb]
\centering
\includegraphics[width=\textwidth]{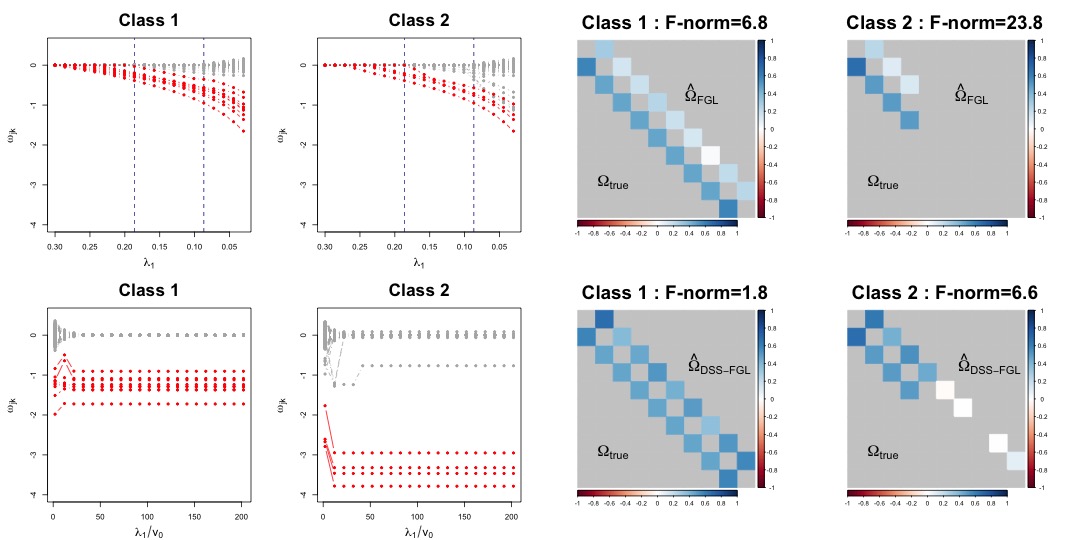}
\caption{The solution paths and estimated precision matrices of FGL (upper row) and DSS-FGL (lower row). 
The red nodes correspond to true edges and the gray nodes correspond to $0$'s. The two vertical lines in the FGL solution path indicate the model that best matches the true sparsity (left) and the model with the lowest AIC (right). 
The block containing the edges is plotted for the estimated values (upper triangular) against the truth (lower triangular). The model that best matches the true graphs is plotted for FGL.
The off-diagonal values are rescaled and negated to partial correlations, and $0$'s are colored with light gray background for easier visual comparison. The bias of the estimated precision matrix as measured by the Frobenius norm, $||\hat\bOmega_g - \bOmega_g||_F$, is also printed in the captions.
 }
\label{fig:path}
\end{figure*}

The algorithm proposed in the previous section requires a fixed set of hyperparameters, $(\lambda_0, \lambda_1, \lambda_2, v_0)$. The posterior is relatively insensitive to the choice of $\lambda_0$ as long as it is not too large~\citep{wang2015scaling}. Furthermore, unlike the original joint graphical lasso, where two tuning parameters need to be selected using cross-validation or model selection criterion, it turns out that we can leverage the self-adaptive property from the doubly spike-and-slab mixture setup to achieve automatic tuning using a path-following strategy~\citep{Veronika2016spike}. Specifically, we consider a sequence of decreasing $v_0 = \{v_0^{1}, ..., v_0^{L}\}$ and some small $\lambda_1$ and $\lambda_2$. 
We initiate $\{\bOmega\}_0$ so that $\bOmega_{g0} = (\bS_g/n_g + c\bm I)^{-1}$, and iterative estimate $\{\widehat \bOmega\}_l$ with $v_0 = v_0^l$. After fitting the $l$-th model, we use the estimated graph structure to warm start the $(l+1)$-th model by initiating $\bOmega_g$ to be $\bOmega_{g0}\circ\bm 1_{\hat\bdelta>0}^{l}$, where $\bm 1_{\hat\bdelta>0}^{l}$ denotes the group level graph structure at the $l$-th iteration. 
As $v_0$ decreases, the shrinkage imposed on the spike elements steadily increases and leads to sparser models. As noted in~\citet{Veronika2016spike}, the solution path from such dynamic reinitialization procedure usually `stabilizes' as $v_0$ becomes closer to $0$ in linear regression. We found similar behavior in our spike-and-slab joint graphical lasso models too, as illustrated in Figure~\ref{fig:path}.

To demonstrate the dynamic posterior exploration in action, we simulated a small dataset from two classes, with $n_g=150$ for $g=1,2$, and $p=100$. The two underlying graphs differ by $5$ edges: The first precision matrix contains a $10$-node block with an AR(1) precision matrix where $(\Omega^{-1})_{jk} = \rho_1^{|j-k|}$, and $\rho_1 = 0.7$; the second precision matrix in the second class contains a common $5$-node AR(1) block with $\rho_2 = 0.9$. The rest of the nodes are all independent. We fit the fused graphical lasso with a sequence of $\lambda_1$, and fixed $\lambda_2 = 0.1$, which leads to the best performance in this experiment; and DSS-FGL with $\lambda_1=1,$ and $\lambda_2 = 1$. Figure~\ref{fig:path} shows the FGL and DSS-FGL solution path. Unlike the continuous shrinkage of FGL, the zero and non-zero elements under DSS-FGL tend to be separated into two stable clusters as the effective shrinkage $\lambda_1/v_0$ increases beyond a critical point. ~\citet{danaher2014joint} noted that graph selection using AIC tends to favor large models. This example also confirms this observation as the likelihood evaluation for smaller models suffers from the overly aggressive shrinkage. In this example, AIC selects $27$ edges in both classes, leading to $41$ false positives. Assuming we know the true graphs, the best model in terms of edge selection along the FGL solution path contains one false negative edge as shown in Figure~\ref{fig:path}. However, without accurate prior knowledge of graph sparsity, correctly identifying this model is typically difficult, if not impossible. On the other hand, the stable model from the DSS-FGL solution path yields $4$ false positive edges in the second graph, but with clear visual separation from the regularization plot: only one false positive edge stabilizing to a larger value away from $0$. Thus in practice, the solution path also provides a visual tool to threshold the small values close to $0$. Additionally, the bias of the final precision matrices compared to the truth is also much smaller than the best FGL solution.

 % are block diagonal with two $10$-node blocks. The first block in the each class are initiated as , and $\rho = (0.7, 0.7, 0.3)$. Then we randomly selected $5$ edges and deleted from the second and the third class. All three classes share the same second block as another AR(1) precision matrix. In another word, the first and second precision matrices differ by $5$ missing edges, and the second and the third precision matrices differ by the values in $5$ of the edges. We fit both the fused graphical lasso with the same $\lambda_1=\lambda_2=\lambda$ for different choices of $\lambda$ and DSS-FGL with $\lambda_1=\lambda_2=1$ and a sequence of $v_0$. 

We also find that the converged region is insensitive to the choice of $\lambda_1$ and $\lambda_2$ in all our experiments, as the model allows a flexible combination of shrinkage through the adaptive estimation of $p^*$. The supplement includes an empirical assessment of sensitivity in the simulation experiments.

\section{Numerical results}
\label{sec:experiment}

\begin{figure*}[tb]
\centering
\includegraphics[width=\textwidth]{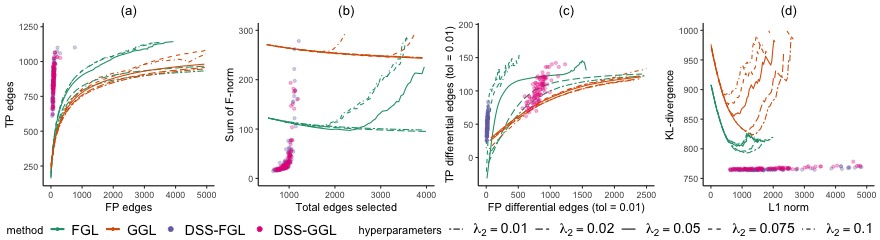}
\caption{Performance of FGL, GGL, DSS-FGL, and DSS-GGL over $100$ replications. 
The dots represent the metrics for the $100$ selected models under DSS-FGL and DSS-GGL, and the lines represent the average performance of  FGL and GGL over $100$ replications under different tuning parameters. 
 }
\label{fig:sim}
\end{figure*}

\paragraph{ Simulation experiments }
\label{sec:sim}
To assess the performance of the proposed models, we consider a three-class problem similar to the study carried out in~\citet{danaher2014joint}. We first generate three networks with $p=500$ features with 10 equal sized unconnected subnetworks following power law degree distributions. Exactly one and two subnetworks are removed from the second and third class. The details of the data generating process can be found in the supplement. The results comparing the proposed model and joint graphical lasso are shown in Figure~\ref{fig:sim}. As discussed before, the DSS-FGL and DSS-GGL achieve model selection automatically. Thus we compare the selected models with the average curve of FGL and GGL under different tuning parameters. Figure~\ref{fig:sim}(a) and (c) show that DSS-FGL and DSS-GGL usually achieves better structure learning performance for both identifying edges and \textit{differential edges}. The differential edges are defined as the edges for the $(g, g')$ pair with $|\omega_{jk}^{(g)}-\omega_{jk}^{(g')}| > 0.01$. Figure~\ref{fig:sim}(b) and (d) demonstrate the bias-diminishing property of the proposed models compared to the joint graphical lasso estimator at varying sparsity levels, measured by the number of edges in (b) and $L_1$ norm of the estimator in (d). On average, both the sum of bias as measured by the Frobenius norm, $||\hat\bOmega_g - \bOmega_g||_F$, and the Kullback-Leibler (KL) divergence achieved by the proposed model is much smaller. 

\paragraph{Symptom networks of verbal autopsy data}
\label{sec:va}
We applied the DSS-FGL and DSS-GGL to a gold-standard dataset of verbal autopsy (VA) surveys~\cite{murray2011population}. VA surveys are widely adopted in countries without full-coverage civil registration and vital statistics systems to estimate cause of death. They are conducted by interviewing caregiver of a recently deceased person about the decedent's health history. The standard procedure of preparing the collected data is to dichotomize all continuous variables into binary indicators and many algorithms have been proposed to automatically assign causes of death using the binary input~\citep{byass2012strengthening,serina2015shortened,insilico}. However, more information may be gained by modeling the continuous variables directly~\citep{li2017mix}. Here we focus on modeling the joint distribution of the continuous variables. The $27$ continuous variables in this dataset contain representations of the duration of symptoms, such as response to the question `how many days did the fever last', and age of the decedents. It is usually reasonable to assume the response to these questions are jointly distributed in similar ways conditional on each cause of death. We take the raw responses and transform raw duration $x_{ij}$ by $\log(x_{ij} + 1)$. We then let $X_{ij}^{(g)}$ denote the $j$-th transformed variable for observation $i$ due to the cause $g$. The full dataset contains death assigned to $34$ causes. We applied DSS-FGL with $\lambda_1=\lambda_2=1$ to the three largest determined causes of death in this data: Stroke ($n=630$), Pneumonia ($n=540$), and AIDS ($n=542$) in Figure~\ref{fig:va}. The estimated graphs under other models are discussed in the supplement. Both DSS-FGL and DSS-GGL estimated similar graphs and discovered interesting differential symptom pairs, such as the strong conditional dependence between the duration of illness and paralysis in deaths due to stroke. Further incorporating the DSS-FGL and DSS-GGL formulation of multiple precision matrices into a classification framework would likely improve accuracy over existing methods (e.g.~\citet{insilico,byass2012strengthening}) for automatic cause-of-death assignment.
\begin{figure*}[tb]
\centering
\includegraphics[width=\textwidth]{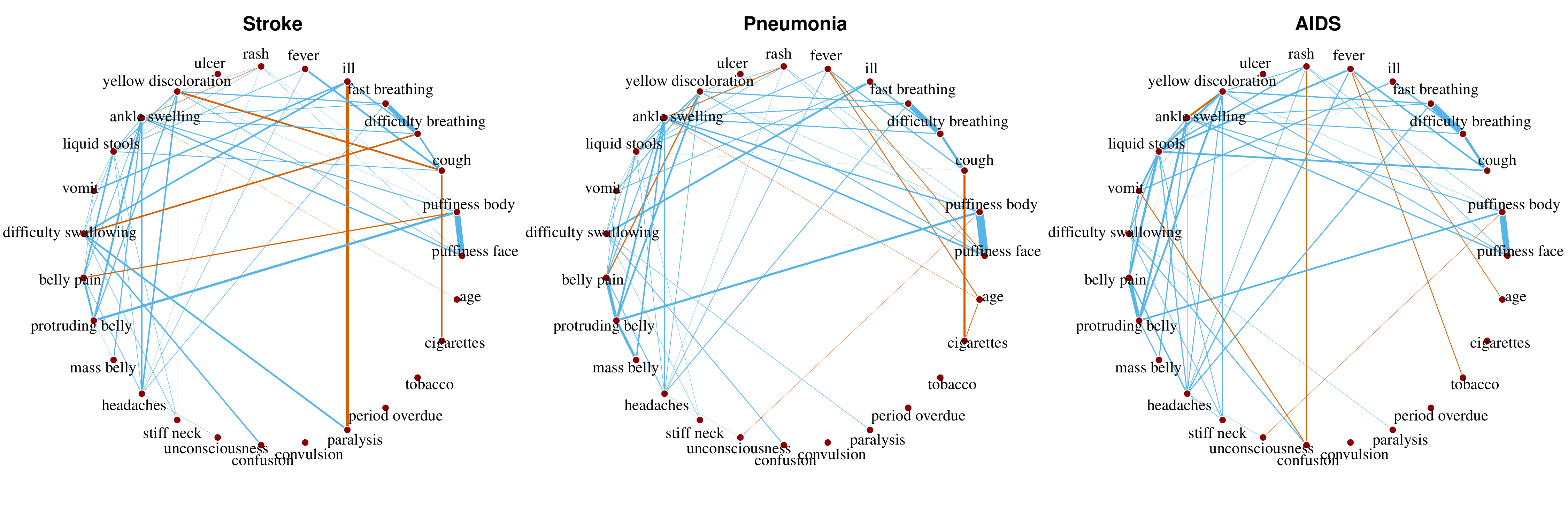}
\vspace{-.5cm}
\caption{Estimated edges between the symptoms under the three causes using DSS-FGL. The width of the edges are proportional to the size of $|\omega_{jk}^{(g)}|$. Common edges across all groups are colored in blue, and the differential edges are colored in red.
 }
\label{fig:va}
\end{figure*}

\paragraph{ Prediction of missing mortality rates}
\label{sec:mortality}
Beyond structure learning, the bias reduction in estimating $\{\bOmega\}$ also makes the proposed method more appealing for prediction tasks involving sparse precision matrices. In this example, we illustrate the potential of using the proposed methods to impute missing mortality rates using a cross-validation study. We construct the data matrices $X_{ij}^{(g)}$ as the log transformed central mortality rate of age group $j$ in year $i$ for subpopulation $g$ (e.g., male and female). Standard approaches in demography, such as the Lee-Carter model~\citep{lee1992modeling}, typically use dimension reduction techniques to estimate mean effects due to age and time, and consider the residuals as independent measurement errors. However, residuals from such models are usually still highly correlated~\citep{fosdick2014separable}. We consider estimating the residual structure with the $1\times 1$ gender-specific mortality table up to age $100$ in the US over the period of 1960 to 2010 using data obtained from the Human Mortality Database (HMD)~\citep{hmd}. For both the male and female mortality, we first randomly selected $25$ years and remove $25$ data points in each of those years.  We then fit a Lee-Carter model to estimate the mean model and interpolate the missing rates. Next, we estimate the covariance matrices among the $101$ age groups in both genders using FGL and DSS-FGL from the residuals. The estimated residuals for the missing values can then be obtained by the E-step in our EM algorithm, or as the expectation from the conditional Gaussian distributions with covariance matrices estimated by FGL. The average mean squared errors (MSEs) for the prediction of missing log rates are summarized in Table~\ref{tab-mort}. Imputation based on DSS-FGL precision matrix reduces the MSE by $27.8\%$ compared to simple interpolation of the mean model (i.e., assuming i.i.d errors), compared to the $6.5\%$ reduction from the FGL precision matrix with the same complexity. The estimated graphs are in the supplement. 
\begin{table}[htb]
  \caption{Average and standard deviation of the mean squared errors from 50 cross-validation experiments. The FGL model is selected to have the same number of edges as the DSS-FGL.}
  \label{tab-mort}
  \centering
  \begin{tabular}{llll}
    \toprule
                 & i.i.d & FGL  & DSS-FGL\\
    \midrule
    Average MSE                   &0.00372&0.00348& 0.00268\\
    SE of the MSEs&0.00030&0.00031& 0.00028\\
    \bottomrule
  \end{tabular}
\end{table}
 \section{Discussion}
 \label{sec:discuss}
 In this paper, we introduced a new class of priors for joint estimation of multiple graphical models. The proposed doubly spike-and-slab mixture priors, DSS-FGL and DSS-GGL, provide self-adaptive extensions to the joint graphical lasso penalties, and achieves simultaneous model selection and parameter estimation. Moreover, while taking advantage of the flexible class of penalty functions, the dynamic posterior exploration procedure allows the penalties to be adaptively estimated in a data-driven way, thus freeing practitioners from choosing multiple tuning parameters. This is especially useful in domains where sample sizes are too small to reliably perform cross-validation. Finally, additional procedures such as multiple random initializations and deterministic annealing may be further incorporated into the proposed algorithm to better explore the posterior surface. While not discussed in the main paper, we note that the posterior uncertainty may be estimated using the Gibbs sampler described in the supplement.
 % , to allow a larger number of models to be explored.    

 The proposed framework can be extended in a few directions. First, we have assumed all classes to be exchangeable, as reflected in the penalty functions for the between-class similarity. When the classes exhibit hierarchical structures or different strengths of similarities, the indicator $\bxi$ may be modeled as functions of the class membership as well. Markov Random Field priors discussed in~\citet{saegusa2016joint} and~\citet{Peterson2014} may also be used to model the between-class similarities. Second, we have considered the estimation of missing values in the data matrices. It is also straightforward to extend to data with missing class labels. In this way, the proposed methods can be extended to classification or discriminant analysis based on sparse precision matrices~\citep{hao2016simultaneous}. Finally, the proposed model is estimated using an EM algorithm that is iteratively solving the joint graphical lasso problem. It may be interesting to construct coordinate ascent algorithms that optimize on the objective function directly, similar to that described in~\citet{Veronika2016spike} for linear regression. 

 The codes for the proposed algorithm are available at \url{https://github.com/richardli/SSJGL}.

\bibliography{MultipleGraph}
\bibliographystyle{icml2019}

\appendix
\onecolumn

%% appendix start here %%

\section{Proof of Proposition~\ref{prop:proper} and~\ref{prop:proper2}}
Here we provide a proof of Proposition~\ref{prop:proper2} in the main paper. The same arguments generalize to Proposition~\ref{prop:proper} directly by fixing all binary indicators to $1$ and thus are not repeated. 

\paragraph{Proof of DSS-GGL prior} We first consider the GGL and DSS-GGL penalties. The joint distribution of all the parameters under the parameterization of the scale mixture of Normal distributions is
\begin{eqnarray*}
p(\{\bOmega\}, \btau, \brho, \bdelta, \bxi, \pi_{\bdelta}, \pi_{\bxi})
&=&\prod_{g}\prod_{j<k}\exp(-\frac{1}{2}(\omega_{jk}^{(g)})^2 (\frac{v_{\delta_{jk}}}{\tau_{jkg}} + \frac{v_{\xi^*_{jk}}}{\rho_{jk}}))
\prod_{g}\prod_{j} \exp(-\frac{\lambda_0}{2}\omega_{jj}^{(g)})
\\  
&&\times \prod_g\prod_{j<k} \tau_{jkg}^{-\frac{1}{2}} \exp(-\frac{\lambda_1^2}{2}\tau_{jkg})  
 \prod_{j<k} \rho_{jk}^{-\frac{1}{2}} \exp(-\frac{\lambda_2^2}{2}\rho_{jk})
 \\
 && \times \prod_{j<k} \pi_{\bdelta}^{\delta_{jk}}(1-\pi_{\bdelta})^{1-\delta_{jk}}  
\prod_{j<k} \pi_{\bxi}^{\xi_{jk}}(1-\pi_{\bxi})^{1-\xi_{jk}}  \\
&& \times \prod_{j<k} \pi_{\bdelta}^{a_1-1}(1-\pi_{\bdelta})^{b_1-1}  
\prod_{j<k} \pi_{\bxi}^{a_2-1}(1-\pi_{\bxi})^{b_2-1}  
\end{eqnarray*}
The following two identities provide the key steps to connect the scale mixture of normal distributions to the Laplace representation in the penalty function:
\begin{eqnarray}
\int_{0}^{\infty} \frac{1}{\sqrt{2\pi s}}\exp(-\frac{z^2}{2s}) \frac{\lambda^2}{2}\exp(-\frac{\lambda^2s}{2}) ds &=& \frac{\lambda}{2}\exp(-\lambda|z|) \\
\int_{0}^{\infty} \frac{1}{\sqrt{2\pi s}}\exp(-\frac{||\bz||_2}{2s}) (\frac{\lambda^2}{2})^{\frac{p+1}{2}}\exp(-\frac{\lambda^2s}{2}) s^{\frac{p+1}{2}-1} \frac{1}{\Gamma(\frac{p+1}{2})}ds &=& \frac{\lambda}{2}\exp(-\lambda||\bz||_2) 
\end{eqnarray}
where $\bz$ is a vector of length $p$ and $||z||_2 = \sqrt{\sum_{i=1}^p z_i^2}$. By rearranging the terms and plugging in the two identities above, it can be seen that
\begin{eqnarray*}
p(\{\bOmega\} | \bdelta, \bxi) &\propto& \exp(-\sum_{j<k} \frac{\lambda_1}{v_{\delta_{jk}}}|\omega_{jk}^{(g)}| - \sum_{j<k}  \frac{\lambda_2}{v_{\xi_{jk}^*}}||\bomega_{jk}||_2-\sum_{g}\sum_{j} \frac{\lambda_0}{2}\omega_{jj}^{(g)}).
\end{eqnarray*}

The conditional distribution of $\{\bOmega\} | \bdelta, \bxi$ takes the form of $\exp(-pen(\{\bOmega\} | \bdelta, \bxi))$ for DSS-GGL, and thus the mode of the posterior is equivalent to the DSS-GGL solution. It still remains to be seen that the intractable constant terms are all finite, so that each of the three conditional distributions are proper. This can be seen as follows:
\begin{eqnarray*}
C_{\btau, \brho}C_{\bdelta, \bxi} &=& \int \prod_{j<k} \mbox{Normal}(\bomega_{jk}; 0, \bTheta) \prod_g\prod_j\mbox{Exp}(\omega_{jj}^{(g)}; \frac{\lambda_0}{2}) \bm 1_{\{\bOmega\} \in M^+}d\{\bOmega\}\\
&<&\int \prod_{j<k} \mbox{Normal}(\bomega_{jk}; 0, \bTheta) \prod_g\prod_j\mbox{Exp}(\omega_{jj}^{(g)}; \frac{\lambda_0}{2}) d\{\bOmega\} = 1\\
C_{\btau, \brho}^{-1} &\propto& \int \prod_{j<k} \big(\exp(-\frac{\lambda_1^2}{2}\sum_g \tau_{jkg} - \frac{\lambda_2^2}{2}\rho_{jk})\rho_{jk}^{-\frac{1}{2}} \prod_g(\tau_{jkg}(\frac{1}{\tau_{jkg}} + \frac{1}{\rho_{jk}}))^{-\frac{1}{2}} \big)d\{\btau_{jk}, \rho_{jk}\} \\
&<&\int \prod_{j<k} \big(\exp(-\frac{\lambda_1^2}{2}\sum_g \tau_{jkg} -\frac{\lambda_2^2}{2}\rho_{jk})\rho_{jk}^{-\frac{1}{2}} \big)d\rho_{jk} \\
&<&\int \prod_{j<k} \big(\exp( - \frac{\lambda_2^2}{2}\rho_{jk})\rho_{jk}^{-\frac{1}{2}} \big)d\rho_{jk} = (2\pi/\lambda_2^2)^{\frac{p(p-1)}{4}}
\end{eqnarray*}
The above inequalities completes the proof that the conditional prior distributions are all proper for DSS-GGL. For GGL prior, the proof is essentially the same by fixing $\bdelta$ and $\bxi$ to be $1$.

\paragraph{Proof of DSS-FGL prior} For DSS-FGL, the joint distribution of all the parameters using the scale Normal mixture representation is
\begin{eqnarray*}
p(\{\bOmega\}, \btau, \brho, \bdelta, \bxi, \pi_{\bdelta}, \pi_{\bxi})
&=&\prod_{g}\prod_{j<k}\exp(-\frac{1}{2}(\bomega_{jk}^{(g)})^T \bTheta_{jk}\bomega_{jk}^{(g)})
\prod_{g}\prod_{j} \exp(-\frac{\lambda_0}{2}\omega_{jj}^{(g)})
\\  
&&\times \prod_g\prod_{j<k} \tau_{jkg}^{-\frac{1}{2}} \exp(-\frac{\lambda_1^2}{2}\tau_{jkg})  
 \prod_{j<k} \rho_{jk}^{-\frac{1}{2}} \exp(-\frac{\lambda_2^2}{2}\rho_{jk})
 \\
 && \times \prod_{j<k} \pi_{\bdelta}^{\delta_{jk}}(1-\pi_{\bdelta})^{1-\delta_{jk}}  
\prod_{j<k} \pi_{\bxi}^{\xi_{jk}}(1-\pi_{\bxi})^{1-\xi_{jk}}  \\
&& \times \prod_{j<k} \pi_{\bdelta}^{a_1-1}(1-\pi_{\bdelta})^{b_1-1}  
\prod_{j<k} \pi_{\bxi}^{a_2-1}(1-\pi_{\bxi})^{b_2-1}  
\end{eqnarray*}
where the first term can be rewritten as the same form as $\exp(-pen(\{\bOmega\} | \bdelta, \bxi))$ for DSS-FGL:
\begin{equation*}
(\bomega_{jk}^{(g)})^T \bTheta_{jk}\bomega_{jk}^{(g)} = \sum_g \frac{v_{\delta_{jk}}}{\tau_{jkg}}(\omega_{jk}^{(g)})^2 + \sum_{g < g'}\frac{v_{\xi_{jk}^*}}{\phi_{jkgg'}}(\omega_{jk}^{(g)} - \omega_{jk}^{(g')})^2 .
\end{equation*}
Using the same identity as before, we can rewrite the conditional distribution below into the form of the DSS-FGL penalty:
\begin{eqnarray*}
p(\{\bOmega\} | \bdelta, \bxi) &\propto& \exp(-\sum_{j<k} \frac{\lambda_1}{v_{\delta_{jk}}}|\omega_{jk}^{(g)}| - \sum_{j<k}\sum_{g<g'}  \frac{\lambda_2}{v_{\xi_{jk}^*}}|\omega_{jk}^{(g)} - \omega_{jk}^{(g')}| - \sum_{g}\sum_{j} \frac{\lambda_0}{2}\omega_{jj}^{(g)}).
\end{eqnarray*}

The proof of the DSS-FGL conditional distributions being proper is similar to the previous case. We 
first note that 
\begin{equation*}
\bTheta_{jk} = \mbox{diag}(\{\frac{1}{\tau_{jkg}}\}_{g=1, ..., G}) + \bm L_{jk}
\end{equation*}
where $\bm L_{jk}$ is a graph Laplacian matrix and is positive semi-definite. Then by Minkowski inequality, $\mbox{det}(\bTheta_{jk}) \geq \mbox{det}(\mbox{diag}(\{\frac{1}{\tau_{jkg}}\}_{g=1, ..., G}))$. Then we have

\begin{eqnarray*}
C_{\btau, \bphi}^{-1} &\propto& \int \prod_{j<k} \big(
\mbox{det}(\bTheta_{jk})^{-\frac{1}{2}}
\exp(-\frac{\lambda_1^2}{2}\sum_g \tau_{jkg} - \frac{\lambda_2^2}{2}\sum_{g < g'}\phi_{jkgg'})
\prod_g \tau_{jkg}^{-\frac{1}{2}}
\prod_{g < g'} \phi_{jkgg'}^{-\frac{1}{2}}
\big)d\{\btau, \bphi\} \\
&\leq& \int \prod_{j<k} \big(
\exp(-\frac{\lambda_1^2}{2}\sum_g \tau_{jkg} - \frac{\lambda_2^2}{2}\sum_{g < g'}\phi_{jkgg'})
\prod_{g < g'} \phi_{jkgg'}^{-\frac{1}{2}}
\big)d\{\btau, \bphi\}\\
&\leq& \int \prod_{j<k} \big(
\exp(- \frac{\lambda_2^2}{2}\sum_{g < g'}\phi_{jkgg'})
\prod_{g < g'} \phi_{jkgg'}^{-\frac{1}{2}}
\big)d\bphi, 
\end{eqnarray*}
which is again finite since the integral consists of products of Gamma densities. The rest of the argument follows in the same way as the DSS-GGL case. 

\section{Details of the EM algorithm implementation}
Assuming no missing data, the full objective function in the $t$-th iteration of the EM algorithm described in~\ref{sec:ecm} is the expectation of the complete data log likelihood, i.e., 
\begin{eqnarray*}
Q(\{\bOmega\}, \pi_{\bdelta}, \pi_{\bxi} | \{\bOmega\}^{(t)}, \pi_{\bdelta}^{(t)}, \pi_{\bxi}^{(t)}) 
&=& E_{\bdelta, \bxi | \{\bOmega\}^{(t)}, \pi_{\bdelta}^{(t)}, \pi_{\bxi}^{(t)}, \bX}
(\log p(\{\bOmega\}, \pi_{\bdelta}, \pi_{\bxi} | \bX)|\{\bOmega\}^{(t)}, \pi_{\bdelta}^{(t)}, \pi_{\bxi}^{(t)}, \bX) \\
&=& \mbox{constant} + \sum_g \frac{n_g}{2}\log|\bOmega_g| - \frac{1}{2}\sum_g tr(\bS_{g}\bOmega_g)
- \frac{\lambda_0}{2}\sum_j\sum_g |\omega_{jj}^{(g)}|
\\
&& - \lambda_1\sum_{j<k}\sum_g |\omega_{jk}^{(g)}|E_{\cdot|\cdot}[\frac{1}{v_0(1-\delta_{jk})+v_1\delta_{jk}}]
+ \sum_{j<k}\log(\frac{\pi_{\bdelta}}{1-\pi_{\bdelta}})E_{\cdot|\cdot}(\delta_{jk})
\\
&& - \lambda_2\sum_{j<k} \widetilde{pen}(\bomega_{jk})E_{\cdot|\cdot}[\frac{1}{v_0(1-\delta_{jk}\xi_{jk})+v_1\delta_{jk}\xi_{jk}}] 
+ \sum_{j<k}\log(\frac{\pi_{\bxi}}{1-\pi_{\bxi}})E_{\cdot|\cdot}(\xi_{jk})
\\
&&  + (a_1-1)\log(\pi_{\bdelta}) + (b_1 + \frac{p(p-1)}{2} - 1) \log(1-\pi_{\bdelta})\\
&& + (a_2-1)\log(\pi_{\bxi}) + (b_2 + \frac{p(p-1)}{2} - 1) \log(1-\pi_{\bxi}),
\end{eqnarray*}
where $E_{\cdot|\cdot}$ denotes conditional expectation $E_{\bdelta, \bxi | \{\bOmega\}^{(t)}, \pi_{\bdelta}^{(t)}, \pi_{\bxi}^{(t)}, \bX}$, and $\widetilde{pen}(\bomega_{jk}) = ||\bomega_{jk}||_2$ for DSS-GGL and $\widetilde{pen}(\bomega_{jk}) = \sum_{g<g'}|\omega_{jk}^{(g)} - \omega_{jk}^{(g')}|$ for DSS-FGL. 

When missing data exists, we need to also calculate the expectation of $\bS_g$ given the observed data. That is, We need to replace the $tr(\bS_g\bOmega_g)$ term in the above objective function by
\begin{equation*}
E_{\cdot|\cdot}(\bS_g\bOmega_g) = E_{\cdot|\cdot}((\frac{1}{n_g}\sum_i^{n_g} \bx_i^{(g)}(\bx_i^{(g)})^T)\bOmega_g) = \frac{1}{n_g}\big(\sum_i^{n_g} E_{\bx^{(g)}_{i,m}|\bx^{(g)}_{i,o}}(\bx_i^{(g)}(\bx_i^{(g)})^T) \big)\bOmega_g.
\end{equation*}
where $\bx^{(g)}_{i,o}$ and $\bx^{(g)}_{i,m}$ denote the observed and missing cells in $\bx^{(g)}_i$ respectively. $\bx^{(g)}_{i}$ follows a multivariate Gaussian distribution. Without loss of generality, if we let $\bx^{(g)}_{i} = \begin{pmatrix}\bx^{(g)}_{i,o}\\\bx^{(g)}_{i,m}\end{pmatrix}$,  we know
 \begin{eqnarray*}
 E_{\bx^{(g)}_{i,m}|\bx^{(g)}_{i,o}}(\bx_{i,m}^{(g)}) &=& 
\bSigma_{mo}\bSigma_{oo}^{-1}\bx^{(g)}_{i,o}
\\
 E_{\bx^{(g)}_{i,m}|\bx^{(g)}_{i,o}}(\bx_i^{(g)}(\bx_i^{(g)})^T) &=&
 E_{\cdot|\cdot}(\bx_i^{(g)}) E_{\cdot|\cdot}(\bx_i^{(g)})^T + 
 \begin{pmatrix}
    \bm 0_{oo}     & \bm 0_{om} \\
    \bm 0_{mo}       & \bSigma_{mm} - \bSigma_{mo}\bSigma_{oo}^{-1}\bSigma_{om}
\end{pmatrix}
 \end{eqnarray*}
 where $\bSigma_{oo}, \bSigma_{om}, \bSigma_{mo}$ and $\bSigma_{mm}$ are the corresponding submatrices of $\bSigma_g$.

\section{Gibbs sampler of the proposed models}
The EM algorithm introduced in the main paper maximizes the complete data likelihood by looking at the Laplace representation after integrating out all the latent parameters. In this section, we show that these latent parameters, $\btau, \bphi$ and $\brho$, facilitates efficient block Gibbs sampling algorithms for fully Bayesian inference.

We start by describing the posterior sampling of $\{\bOmega\}$. The basic idea is to sample each column and row for all the precision matrices jointly. To similify notation, we separate out the last column and row in $\bOmega_g$ and $\bS_g$ and define
\begin{equation*}
\bOmega_g =  \begin{pmatrix}
    \bOmega_{11}^{(g)}      & \bomega_{12}^{(g)}   \\
    \bomega_{21}^{(g)}        & \omega_{22}^{(g)}   
\end{pmatrix},
\;\;\;\;
\bS_g =  \begin{pmatrix}
    \bS_{11}^{(g)}      & \bs_{12}^{(g)}   \\
    \bs_{21}^{(g)}        & s_{22}^{(g)}   
\end{pmatrix}.
\end{equation*}
We further let $\bomega_{12}= [(\bomega_{12}^{(1)})^T, (\bomega_{12}^{(2)})^T, ..., (\bomega_{12}^{(G)})^T]^T$, and $\bs_{12} = [(\bs_{12}^{(1)})^T , (\bs_{12}^{(2)})^T , ..., (\bs_{12}^{(G)})^T]^T$, each denoting a vector of length $(p-1)G$, and $\bomega_{22} = [\omega_{22}^{(1)}, \omega_{22}^{(2)}, ..., \omega_{22}^{(G)}]$.

The conditional distribution of $(\bomega_{12}, \bomega_{22})$ given the rest of the elements in $\{\bOmega\}$ does not seem to take any standard form. However, if we perform a change of variables and let $\theta_g = \omega_{22}^{(g)} - \bomega_{21}^{(g)} (\bOmega_{11}^{(g)})^{-1} \bomega_{12}^{(g)}$, the conditional distribution of $(\bomega_{12}, \btheta)$ becomes 
\begin{eqnarray*}
p(\bomega_{12}, \btheta) &\propto& 
\sum_g \theta_g^{\frac{n_g}{2}} 
\exp(-\frac{s_{22}^{(g)} + \lambda_0}{2}\theta_g - \bs_{12}^T\bomega_{12} - \frac{1}{2}\bomega_{12}^T\bA\bomega_{12}) \\
&=&
\prod_g \mbox{Gamma}(\theta_g; \frac{n_g}{2}+1, \frac{s_{22}^{(g)} + \lambda_0}{2}) \times
\mbox{Normal}(\bomega_{12}; -\bA^{-1}\bs_{12}, \bA^{-1}).
\end{eqnarray*}
The $\bA$ matrix can be calculated by $\bA = \bU + \bV$, where $\bU$ is a matrix by rearrangeing the precision matrices so that its 
$((g-1)(p-1)+k, (g'-1)(p-1)+k)$-th element is the $(g,g')$-element in
$\tilde\bTheta_{jk}$ defined in (\ref{eqn:theta-ssfgl}) and (\ref{eqn:theta-ssggl}) of the main paper, and 
\begin{equation*}
\bV = \begin{pmatrix}
    (\lambda_0 + s_{22}^{(1)})(\bOmega_{11}^{(1)})^{-1}     & &\\
          & \ddots & \\
          & & (\lambda_0 + s_{22}^{(G)})(\bOmega_{11}^{(G)})^{-1}  
\end{pmatrix},
\end{equation*}

For DSS-GGL, we notice that $\bA$ is block diagonal, thus we can alternatively sample $\bomega_{12}^{(g)}$ independently by
\begin{equation*}
\bomega_{12}^{(g)}|\cdot \sim \mbox{Normal}(-\bA_g^{-1}\bs_{12}^{(g)}, \bA_g^{-1})
\end{equation*}
where $\bA_g = (\lambda_0 + s_{22}^{(g)})(\bOmega_{11}^{(g)})^{-1} + \tilde\bTheta_{11}^{(g)}$. 

Given $\{\bOmega\}$, the latent parameters in DSS-GGL have simple conditional distribution as follows:
\begin{eqnarray*}
\tau_{jkg}^{-1} | \cdot &\sim& \mbox{InvGaussian}(\frac{\lambda_1}{v_{\delta_{jk}}^{\frac{1}{2}}|\omega_{jk}^{(g)}|}, \lambda_1^2), \;\;\; j, k = 1,...,p, g = 1,..,G\\
\rho_{jk}^{-1} | \cdot &\sim& \mbox{InvGaussian}(\frac{\lambda_2}{v_{\xi^*_{jk}}\sum_g(\omega_{jk}^{(g)})^2}, \lambda_2^2), \;\;\; j, k = 1,...,p \\
\delta_{jk}, \xi_{jk} | \cdot &\sim& p^*(\bdelta_{jk}, \bxi_{jk}), \;\;\; j, k = 1,...,p  \\
\pi_{\bdelta} &\sim& \mbox{Beta}(\sum \delta_{jk} + a_1, \sum (1-\delta_{jk}) + b_1) \\
\pi_{\bxi} &\sim& \mbox{Beta}(\sum \xi_{jk} + a_2, \sum (1-\xi_{jk}) + b_2) \end{eqnarray*}
where $p^*(\bdelta, \bxi)$ is defined in (\ref{eqn:pstar}) in the main paper. 

For DSS-FGL, the conditional distribution of $\btau, \bdelta, \bxi, \pi_{\bdelta}$, and $\pi_{\bxi}$ are the same as DSS-GGL. The conditional distribution of $\bphi$ is
\begin{equation*}
\phi^{-1}_{jkgg} \sim \mbox{InvGaussian}(\frac{\lambda_2}{v_{\xi^*_{jk}}^{\frac{1}{2}}|\omega_{jk}^{(g)}-\omega_{jk}^{(g')}|}, \lambda_2^2), \;\;\; j, k = 1,...,p, g, g' = 1,..,G
\end{equation*}
The Gibbs sampler is then complete by circling through and sampling each blocks of $\{\bOmega\}$ and the latent parameters with the above posterior conditional distributions. 

\section{Additional illustrating example}
In this section, we provide a more detailed description to the small $2$-class simulated example described in Section~\ref{sec:selection} of the main paper. We let $n_g=150$ for $g=1,2$, and $p=100$. The first $10$ variables in the first class form a $10$-node block with an AR(1) precision matrix, i.e., $(\bOmega^{-1})_{jk} = 0.7^{|j-k|}$. The rest of the $90$ variables are independent noises from a standard Gaussian distribution. The second class shares the first $4$ edges of the first class. The first $5$ variables form another AR(1) block with different strength of correlations so that $(\bOmega^{-1})_{jk} = 0.9^{|j-k|}$. The rest of the $95$ variables all follow independent standard Gaussian distribution. 

The best performance of FGL in our experiments was achieved with $\lambda_2 = 0.1$. We then obtained the regularization path along different values of $\lambda_1$. We also fit DSS-FGL and SS-FGL with $\lambda_1=1, \lambda_2 = 1$. The latter assumes $\xi_{jk}=1$ for all edges, i.e., the penalization of similarities is always proportional to the penalization of sparsity.

\begin{figure*}[tb]
\centering
\includegraphics[width=\textwidth]{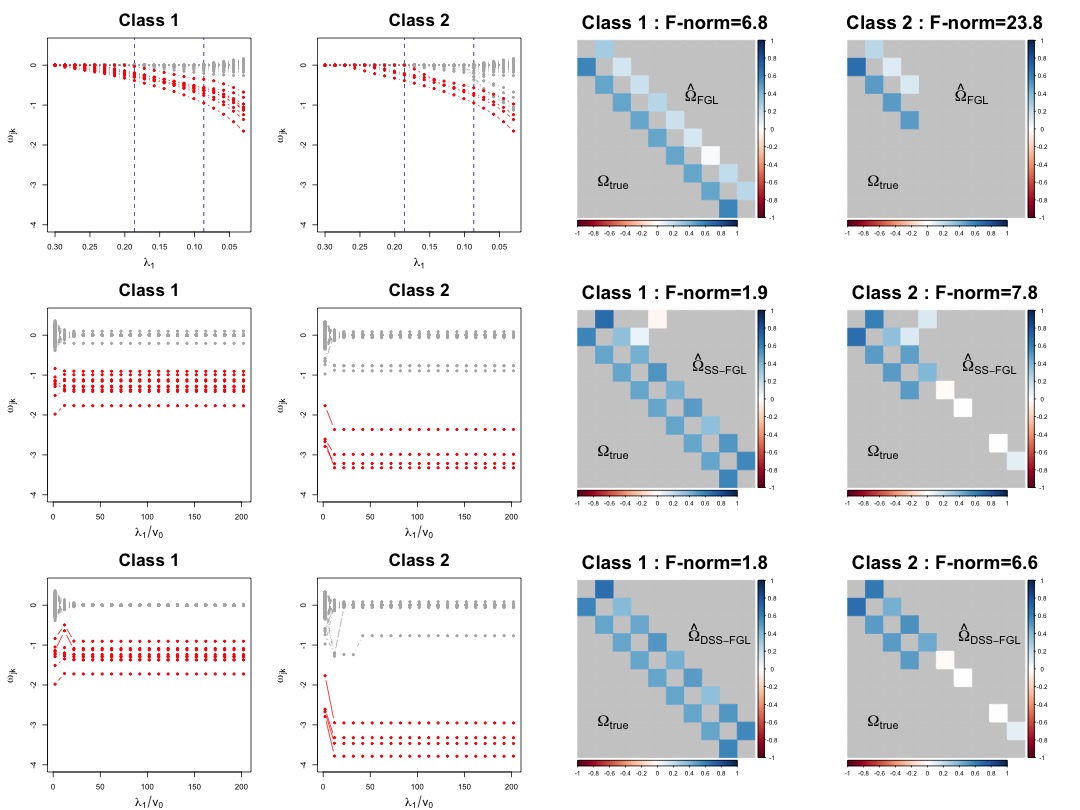}
\caption{The solution paths and estimated precision matrices of FGL (upper row), SS-FGL (middle row) and DSS-FGL (lower row). 
The red nodes correspond to true edges and the gray nodes correspond to $0$'s. The two vertical lines in the FGL solution path indicate the model that best matches the true sparsity (left) and the model with the lowest AIC (right). 
The block containing the edges is plotted for the estimated values (upper triangular) against the truth (lower triangular). The model that best matches the true graphs is plotted for FGL.
The off-diagonal values are rescaled and negated to partial correlations, and $0$'s are colored with light gray background for easier visual comparison. The bias of the estimated precision matrix measured by the Frobenius norm, $||\hat\bOmega_g - \bOmega_g||_F$, is also printed in the captions.
 }
\label{fig:path2}
\end{figure*}

Figure~\ref{fig:path2} shows both the two solution paths presented in the main paper, and the solution path from SS-FGL. It can be seen that although SS-FGL achieves similar bias as DSS-FGL, it also estimates several more false positive edges. This can be seen from the formulation of the doubly spike-and-slab selection: with only one spike-and-slab mixture of the penalties, the selected edges from the slab distributions receive also only weak penalization for between-class similarities. Thus it is more likely to pick up spurious edges due to noises that happen to exist in one class. This full illustration of the simple example shows the advantage of having the doubly spike-and-slab setup.

\section{Additional simulation evidence}
\begin{figure*}[tb]
\centering
\includegraphics[width=.8\textwidth]{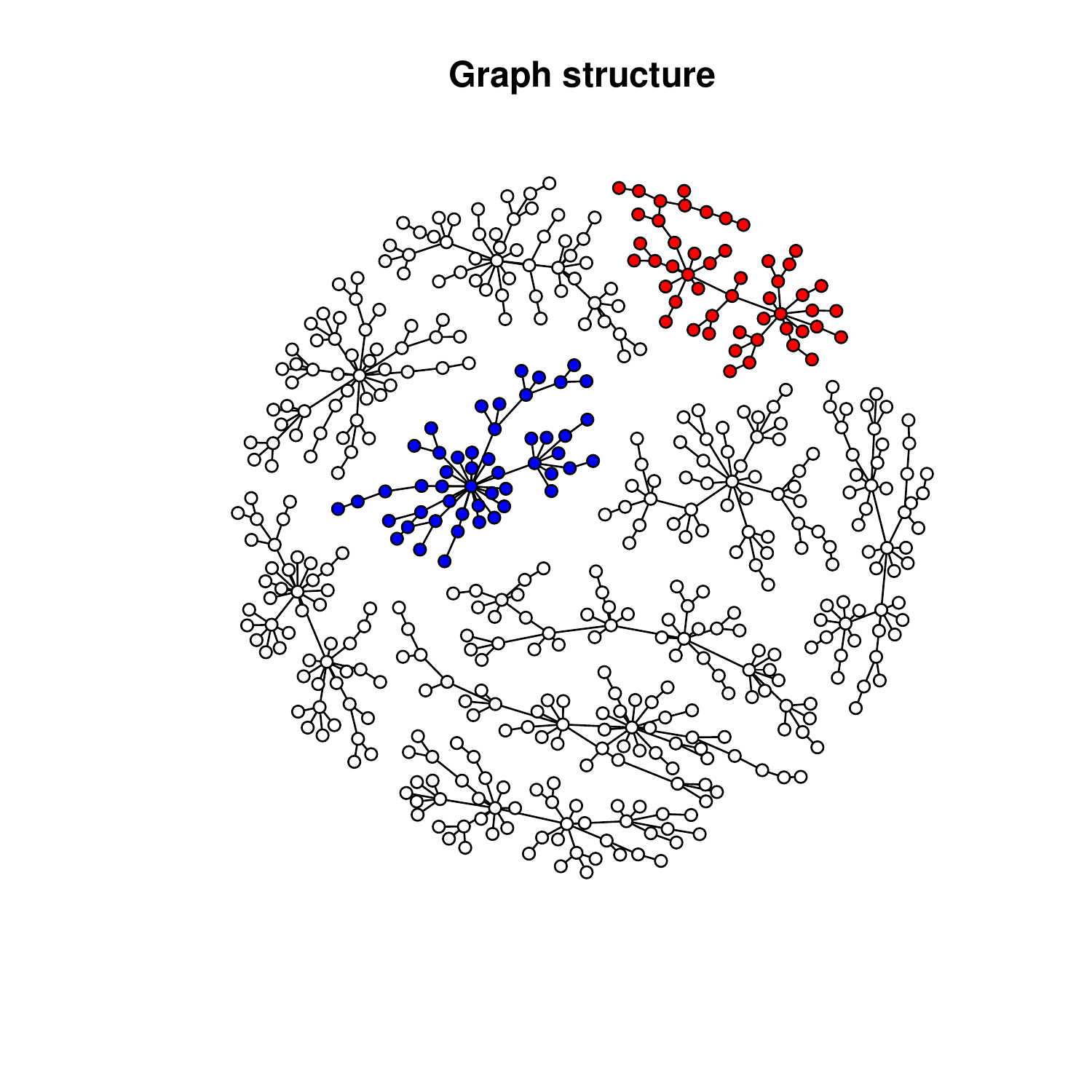}
\vspace{-2cm}
\caption{Graph structure of the simulated dataset. The edges between the red nodes are removed from the second class, and edges between both the red and blue nodes are removed from the third class.}
\label{fig:sim2}
\end{figure*}

Here we describe our procedure in simulating the dataset described in Section~\ref{sec:sim} of the main paper. We first generate three networks with $p$ features with 10 equal sized unconnected subnetwork. Each of the subnetwork follow a power law degree distribution, which is generally harder to estimate than simpler structures~\citep{peng2009partial}. The first class contains all ten subnetworks, and the second and third classes each has one and two subnetworks removed. Given the network structure, we generate $\bOmega_g$ from the $G$-Wishart distribution $W_G(3, I_p)$, and rescale them so that $\bOmega_g^{-1}$ have unit variances. Finally, we generate $n = 150$ independent and identically distributed samples from $\mbox{Normal}(\bm 0, \bOmega_g)$ in each class. The resulted graph for $p=500$ is shown in Figure~\ref{fig:sim2}. We fit GGL and FGL with various choice of fixed $\lambda_2$ and a sequence of $\lambda_1$. We fit DSS-GGL and DSS-FGL with $\lambda_1 = 1$, $\lambda_2 = 2/30$ in this case. We explore more choices of $\lambda_2$ in the moderate dimensional experiments below and found no substantial changes in the performance of the final models.

\begin{figure*}[tb]
\centering
\includegraphics[width=\textwidth]{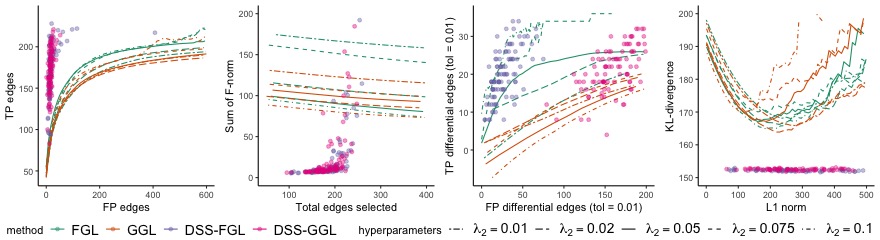}
\caption{Performance of FGL, GGL, DSS-FGL, and DSS-GGL over $100$ replications, $p = 100$. 
The dots represent the metrics for the $100$ selected models under DSS-FGL and DSS-GGL, and the lines represent the average performance of  FGL and GGL over $100$ replications under different tuning parameters. 
 }
\label{fig:sim3}
\end{figure*}

\begin{figure*}[tb]
\centering
\includegraphics[width=\textwidth]{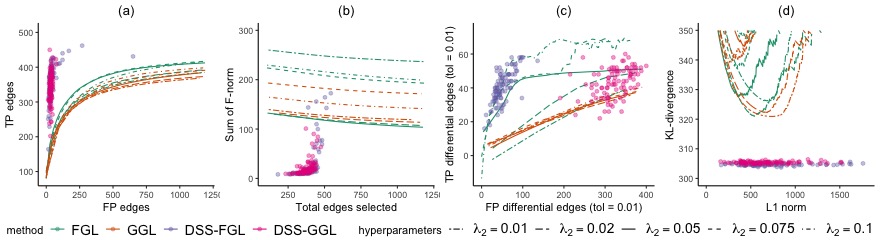}
\caption{Performance of FGL, GGL, DSS-FGL, and DSS-GGL over $100$ replications, $p=200$. 
The dots represent the metrics for the $100$ selected models under DSS-FGL and DSS-GGL, and the lines represent the average performance of  FGL and GGL over $100$ replications under different tuning parameters. 
 }
\label{fig:sim4}
\end{figure*}

Additional results for $p=100$ and $p=200$ are shown in Figure~\ref{fig:sim3} and \ref{fig:sim4}. The dots correspond to DSS-GGL and DSS-FGL with $\lambda_1=1$, $\lambda_2=0.1$. We also examined different choices of $\lambda_2$ are found no substantial differences in performance. An exploratory sensitivity analysis is presented in the next subsection.

\subsection{Sensitivity to hyperparameters}
\begin{figure*}[tb]
\centering
\includegraphics[width=\textwidth]{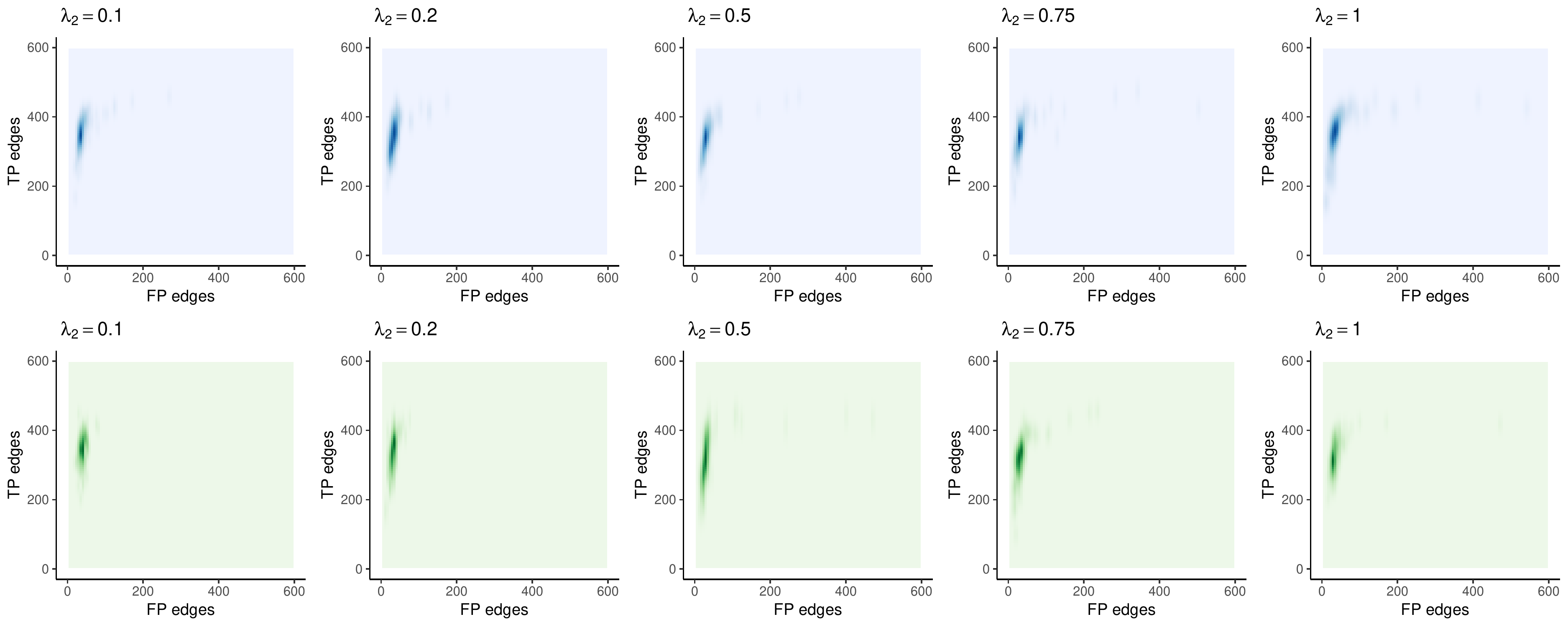}
\caption{The density plot of true positive edges against false positive edges for DSS-FGL (top row), and DSS-GGL (bottom row) under different choices of $\lambda_2$. $\lambda_1$ is set to $1$.}
\label{fig:sens1}
\end{figure*}

\begin{figure*}[tb]
\centering
\includegraphics[width=\textwidth]{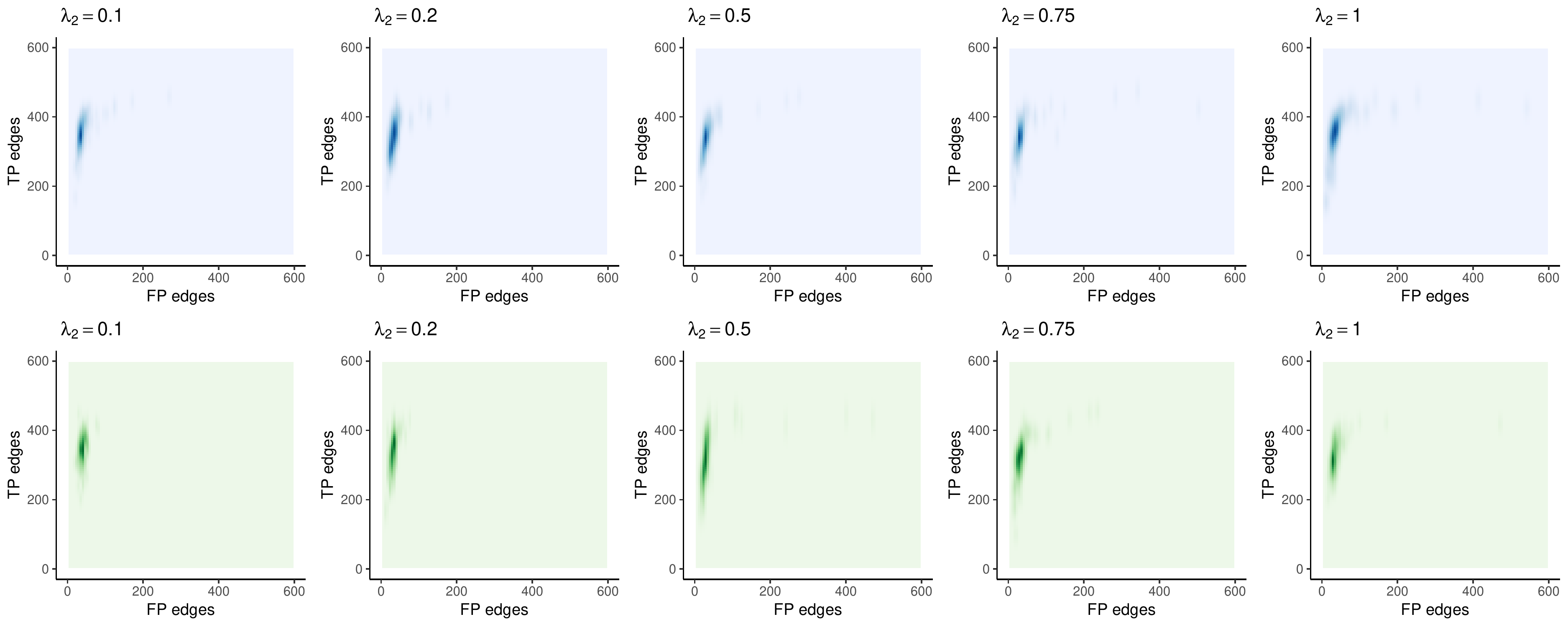}
\caption{The density plot of true positive differential edges against false positive positive edges for DSS-FGL (top row), and DSS-GGL (bottom row) under different choices of $\lambda_2$. $\lambda_1$ is set to $1$.}
\label{fig:sens2}
\end{figure*}

Figure~\ref{fig:sens1} and \ref{fig:sens2} shows the converged regions over $100$ replications on space of the true positive against false positive discoveries for edges and differential edges respectively when $p=200$. It can be seen that the performance is relatively stable under different choices of $\lambda_2$.

\clearpage
\section{Details on the verbal autopsy data analysis}
 \subsection{List of symptoms}
 Table~\ref{tab:va} shows the questions with continuous responses used in the analysis in the main paper. 
\begin{table}[hb]
  \caption{List of symptoms considered in this analysis.}
  \label{tab:va}
  \centering
  \begin{tabular}{llll}
    \toprule
    Abbreviation & Questionnaire item \\
    \midrule
    ill& For how long was [name] ill before s/he died? [days]\\
    fever&  How many days did the fever last? [days]\\
    rash&  How many days did [name] have the rash? [days]\\
    ulcer&  For how many days did the ulcer ooze pus? [days]\\
    yellow discoloration &  For how long did [name] have the yellow discoloration? [days]\\
    ankle swelling &  For how long did [name] have ankle swelling? [days]\\
    puffiness face&  For how long did [name] have puffiness of the face? [days]\\
    puffiness body &  For how long did [name] have puffiness all over his/her body? [days]\\
    cough &  For how long did [name] have a cough? [days]\\
    difficulty breathing &  For how long did [name] have difficulty breathing? [days]\\
    fast breathing &   For how long did [name] have fast breathing? [days]\\
    liquid stool &   For how long before death did [name] have loose or liquid stools? [days]\\
    vomit &  For how long before death did [name] vomit? [days]\\
     difficulty swallowing & For how long before death did [name] have difficulty swallowing? [days]\\
    belly pain &  For how long before death did [name] have belly pain? [days]\\
    protruding belly &  For how long before death did [name] have a protruding belly? [days]\\
    mass belly &  For how long before death did [name] have a mass in the belly [days]\\
    headaches &  For how long before death did [name] have headaches? [days]\\
    stiff neck &  For how long before death did [name] have stiff neck? [days]\\
    unconsciousness &  For how long did the period of loss of consciousness last? [days]\\
    confusion &  For how long did the period of confusion last? [days]\\
    convulsion &  For how long before death did the convulsions last? [days]\\
    paralysis &  For how long before death did [name] have paralysis? [days]\\
    period overdue &  For how many weeks was her period overdue? [days]\\
    tobacco &   How much pipe/chewing tobacco did [name] use daily?\\
    cigarettes &  How many cigarettes did [name] smoke daily?\\
    age &  Age [years]\\  
    \bottomrule
  \end{tabular}
\end{table}
\subsection{Comparing with JGL}
\begin{figure*}[tb]
\centering
\includegraphics[width=\textwidth]{figures/phmrc1.pdf}
\includegraphics[width=\textwidth]{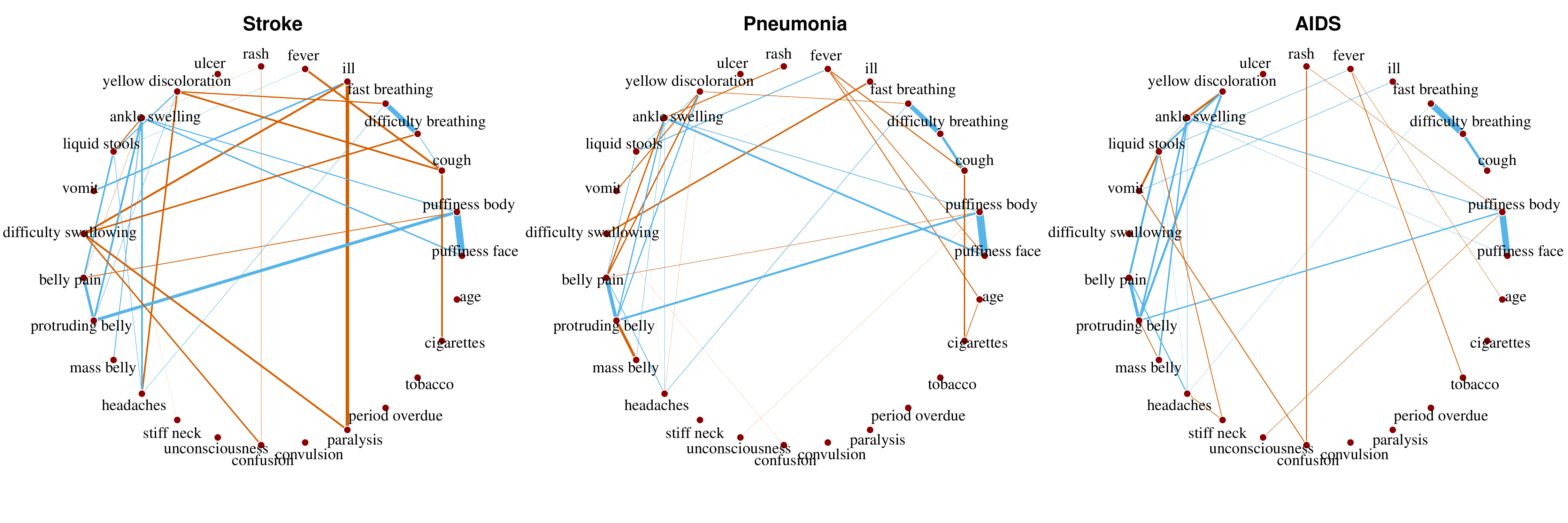}
\caption{Estimated edges between the symptoms under the three causes using DSS-FGL (top row) and DSS-GGL (bottom row). The width of the edges are proportional to the size of $|\omega_{jk}^{(g)}|$. Common edges across all groups are colored in blue, and the differential edges are colored in red.
 }
\label{fig:va1}
\end{figure*}

The estimated symptom network from the DSS-FGL and DSS-GGL are summarized in Figure~\ref{fig:va1}.

\begin{figure*}[tb]
\centering
\includegraphics[width=\textwidth]{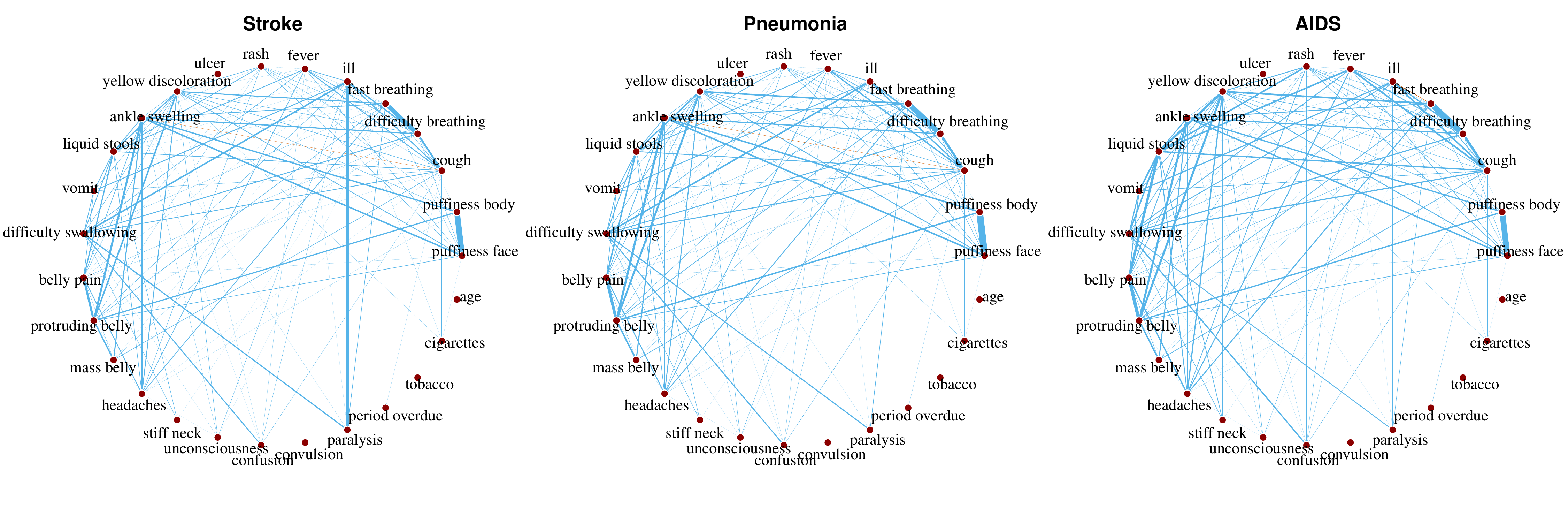}
\includegraphics[width=\textwidth]{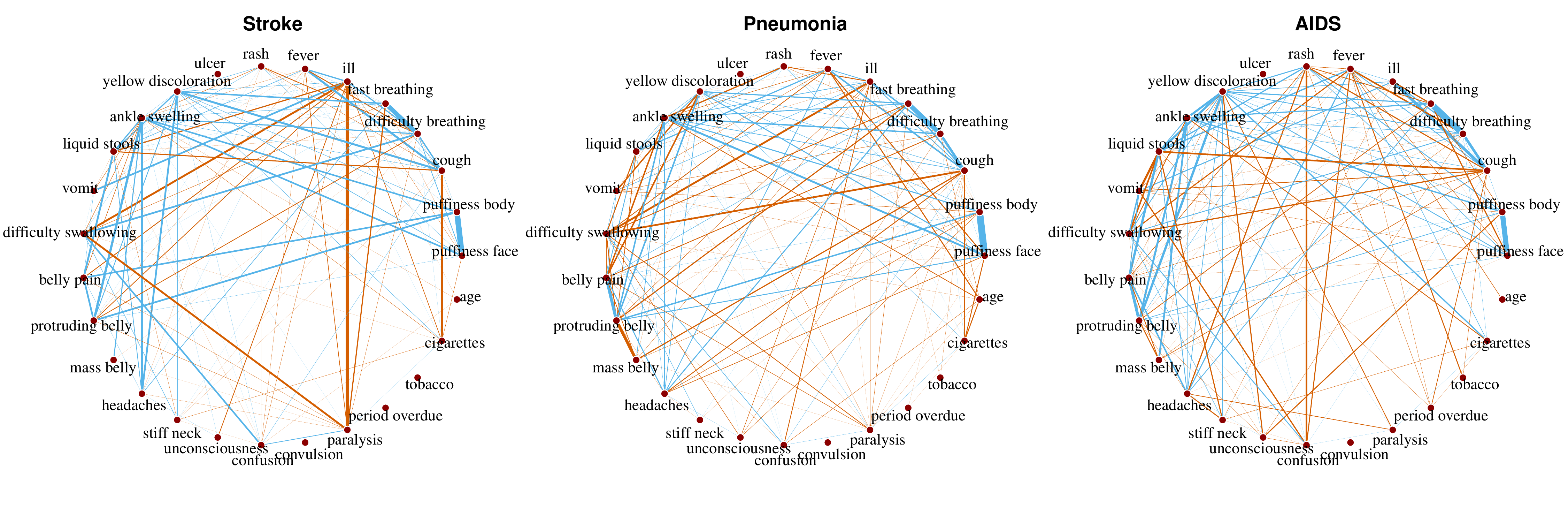}
\includegraphics[width=\textwidth]{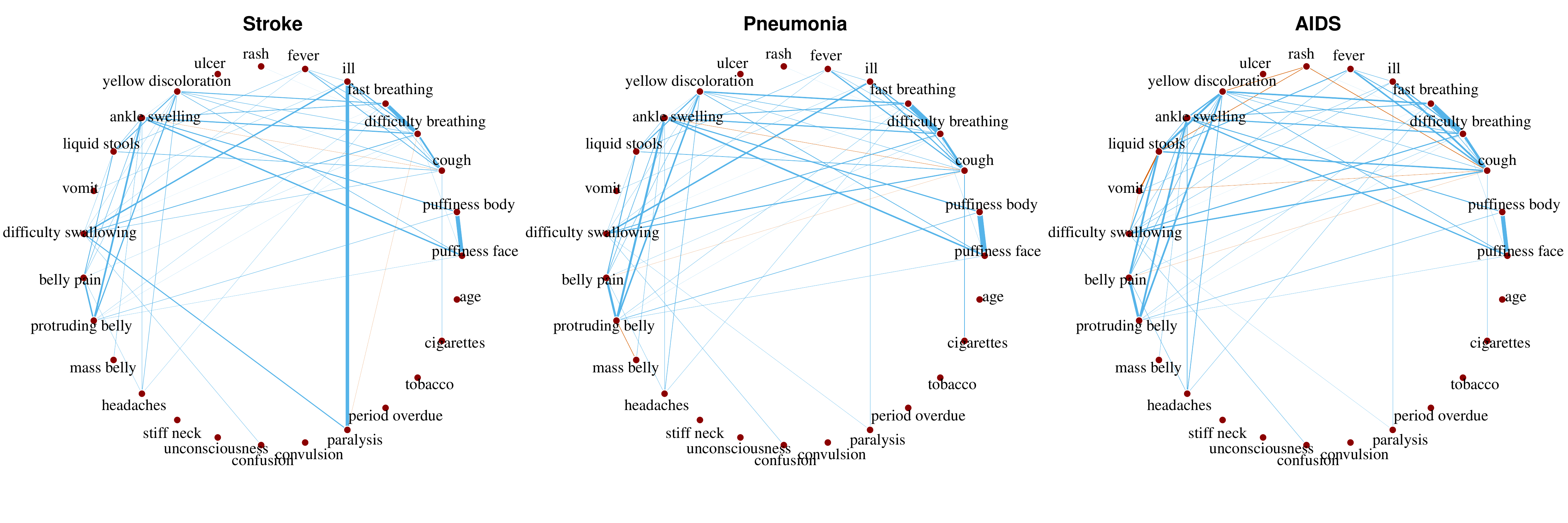}
\includegraphics[width=\textwidth]{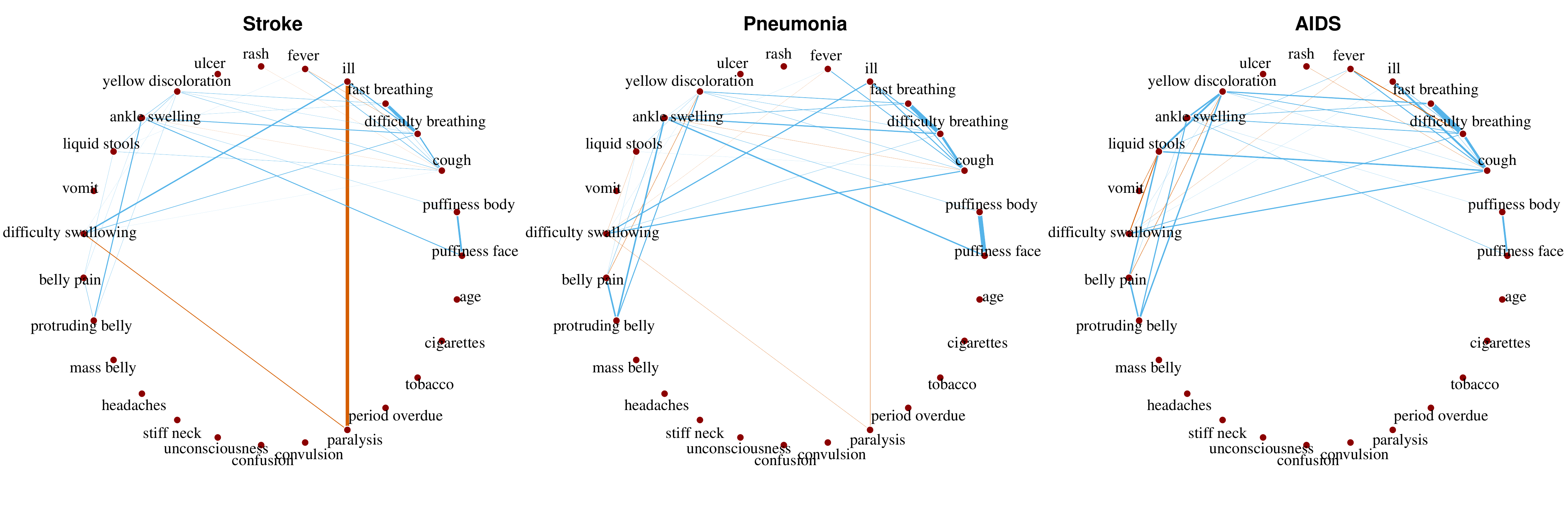}
\caption{Estimated edges between the symptoms under the three causes using FGL using AIC (first row), GGL using AIC (second row), FGL with the same number of edges as selected by DSS-FGL (third row), and GGL with the same number of edges as selected by DSS-GGL (last row).  The width of the edges are proportional to the size of $|\omega_{jk}^{(g)}|$. Common edges across all groups are colored in blue, and the differential edges are colored in red.
 }
\label{fig:va2}
\end{figure*}

The estimated symptom network from the FGL and GGL are summarized in Figure~\ref{fig:va2}. We fit both models under a 2-dimensional grids over $\lambda_1$ and $\lambda_2$. As expected, AIC selects very dense graphs (first two rows of Figure~\ref{fig:va2}) and are difficult to interpret. We also compare the FGL and GGL graph with the closest number of edges as those from DSS-FGL and DSS-GGL in the third and forth row of Figure~\ref{fig:va2}. The number of differential edges is typically smaller compared to DSS-FGL and DSS-GGL, which is likely due to over penalization of similarities, i.e., edges become too similar using FGL, and too sparse among half of the nodes using GGL. 

\section{Details on prediction of missing mortality rates}
\begin{figure*}[tb]
\centering
\includegraphics[width=\textwidth]{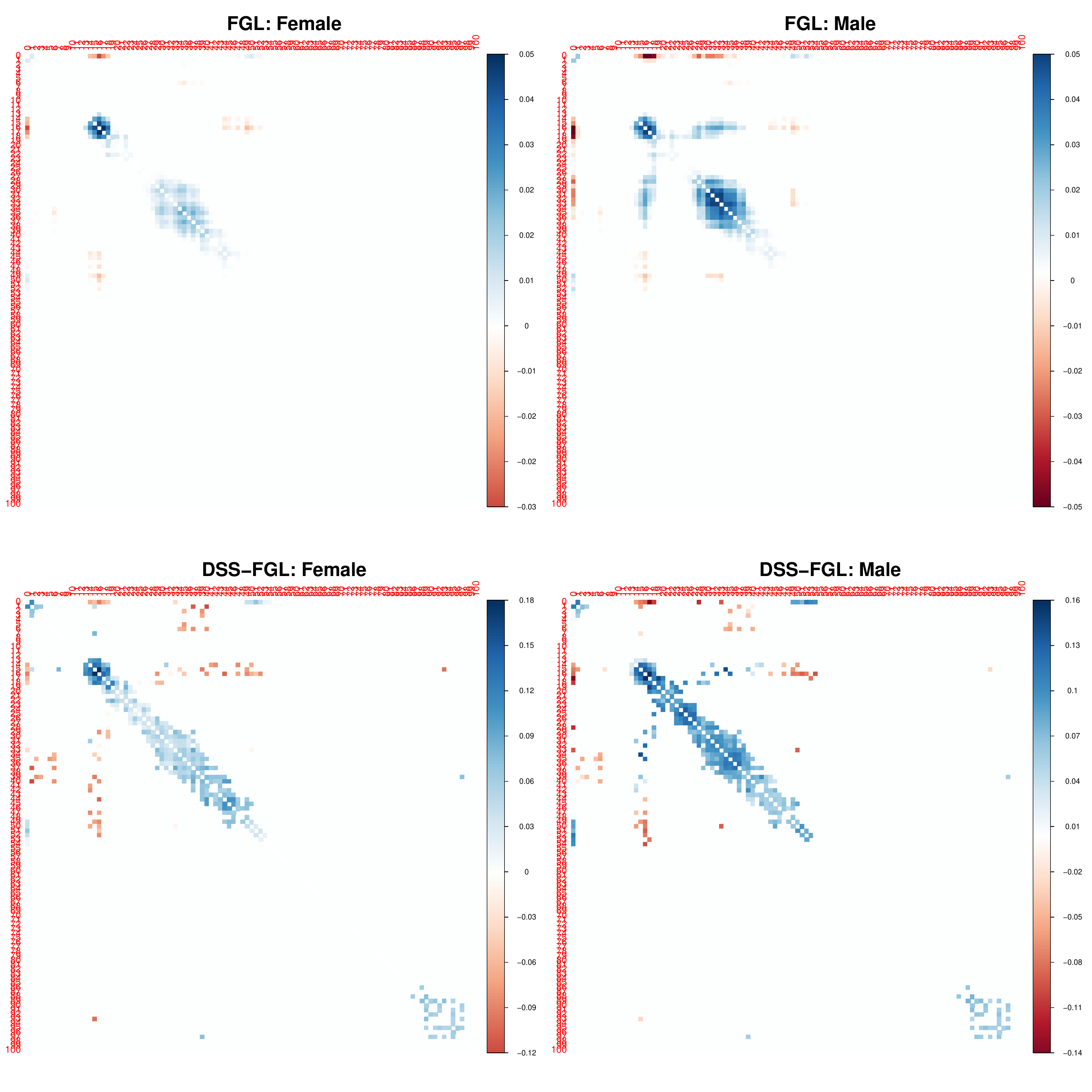}
\caption{Estimated partial correlation matrix using one cross-validation dataset. The partial correlations among the $101$ age groups are estimated using FGL with the same number of edges as selected by DSS-FGL (top row), and DSS-FGL (bottom row). DSS-FGL estimates $197$ and $199$ edges respectively for female and male. The closet configuration of FGL estimates $157$ and $241$ edges respectively. The precision matrices are rescaled and negated to partial correlations for easier interpretation.
 }
\label{fig:mortality2}
\end{figure*}

The data we consider in this example consist of log mortality rates over $n=51$ years for $p=101$ age groups, and $2$ classes representing female and male series respectively. The estimated graph structure from one of the cross-validation dataset using FGL and DSS-FGL are shown in Figure~\ref{fig:mortality2}. The Lee-Carter model are estimated using the R package {\tt ilc}~\citep{ilc} for each gender separately. 

The DSS-FGL is able to pick up more conditional dependence structures along the diagonal among several age groups, while the FGL estimates mostly within adults only. It is interesting that both approaches identifies positive partial correlations between age $14-17$ and $30-40$ between male and female. This is likely due to the fact that male mortality around age $20$ typically shows a hump of increase due to young adult accident mortality, which leads to the mean model more likely to underestimations for mortality during age $18-30$ and overestimations both before and after that period. This relationship of the age curve, however, is not seen in female mortality.

\end{document}